\documentclass[10pt,twocolumn,letterpaper]{article}

\usepackage{iccv}
\usepackage{times}
\usepackage{epsfig}
\usepackage{graphicx}
\usepackage{amsmath}
\usepackage{amssymb}
\usepackage{booktabs}
\usepackage{epigraph} 
\usepackage{multirow}
\usepackage{comment}
\usepackage{float}
\usepackage{tabu}
\usepackage{subfigure}
\usepackage{pifont}
\newcommand{\cmark}{\ding{51}}%
\newcommand{\xmark}{\ding{55}}%
\usepackage[dvipsnames,table]{xcolor}
\definecolor{pastelgray}{rgb}{0.91, 0.91, 0.91}
\usepackage{enumitem}
\usepackage[hang,flushmargin]{footmisc}
\usepackage[numbers,sort]{natbib}

\usepackage[pagebackref=true,breaklinks=true,letterpaper=true,colorlinks,bookmarks=false]{hyperref}

\iccvfinalcopy 
\def\iccvPaperID{3138} 

\newcommand\blfootnote[1]{%
  \begingroup
  \renewcommand\thefootnote{}\footnote{#1}%
  \addtocounter{footnote}{-1}%
  \endgroup
}

\newcommand{\app}[0]{{Appendix}}

\begin{document}


\title{AutoAD II: The Sequel -- Who, When, and What in Movie Audio Description}

\author{
Tengda Han$^{1}$ \quad Max Bain$^{1}$ \quad
Arsha Nagrani$^{1\dagger}$ \quad G\"ul Varol$^{1,2}$ \quad Weidi Xie$^{1,3}$ \quad Andrew Zisserman$^1$\\
{\small$^1$Visual Geometry Group, University of Oxford} \quad
{\small$^2$LIGM, \'Ecole des Ponts ParisTech}
\quad{\small$^3$CMIC, Shanghai Jiao Tong University}
}


\maketitle

\begin{abstract}
Audio Description (AD) is the task of generating descriptions of visual content, at suitable time intervals, for the benefit of visually impaired audiences. For movies, this presents notable challenges -- AD must occur only during existing pauses in dialogue, should refer to characters by name, and ought to aid understanding of the storyline as a whole.

To this end, we develop a new model for automatically generating movie AD, given CLIP visual features of the frames, the cast list, and the temporal locations of the speech; addressing all three of the `who', `when', and `what' questions:
(i) who --
we introduce a {\em character bank} consisting of the character's name, the actor that played the part, and
a CLIP feature of their face, for the principal cast of each movie, and demonstrate how this can be used to improve
naming in the generated AD; (ii) when --
we investigate several models for determining whether an AD should be generated for a time interval or not, based
on the visual content of the interval and its neighbours; and (iii) what --
we implement a new vision-language model for this task, that can ingest the proposals
from the character bank, whilst conditioning on the visual features using cross-attention, and demonstrate how this improves over previous architectures for AD text generation in an apples-to-apples comparison.

\end{abstract}
\blfootnote{$\dagger$: also at Google Research}
\vspace{-3mm}
\section{Introduction}
\label{sec:intro}
\renewcommand{\epigraphflush}{flushleft}
\renewcommand{\epigraphsize}{\small}
\setlength{\epigraphwidth}{0.45\textwidth}
{\scriptsize
\epigraph{\textit{For in acts we must take note of \textbf{who} did it, by what aids or instruments he did it (with), \textbf{what} he did, where he did it, why he did it, how and \textbf{when} he did it.}\hspace{30pt}\textit{Thomas Aquinas}}
{}
}\vspace{-3mm}

Audio Description (AD) is the descriptive narration of visual elements in a video, that are not represented in the original audio track. While there has been a proliferation of online content with \textit{closed captioning}\footnote{Transcription of the speech} due to advancements in ASR, a
vast majority of video online does not have AD, mostly due to the prohibitive cost of generating it (\$30 per minute\footnote{https://www.3playmedia.com/blog/select-audio-description-vendor/}). Generating AD automatically at scale has multiple benefits; not only does it improve access for the visually impaired -- it may also enhance the visual experience for sighted users (sight-free multitasking such as driving, enhanced memory for visual details, language learning, and also aiding those with other cognitive disabilities)~\cite{perego2016gains}. Generating AD for movies is also an important research area in computer vision as it requires a system to perform multi-modal reasoning of long videos over time. 

\begin{figure}[t]
    \centering
    \includegraphics[width=0.5\textwidth]{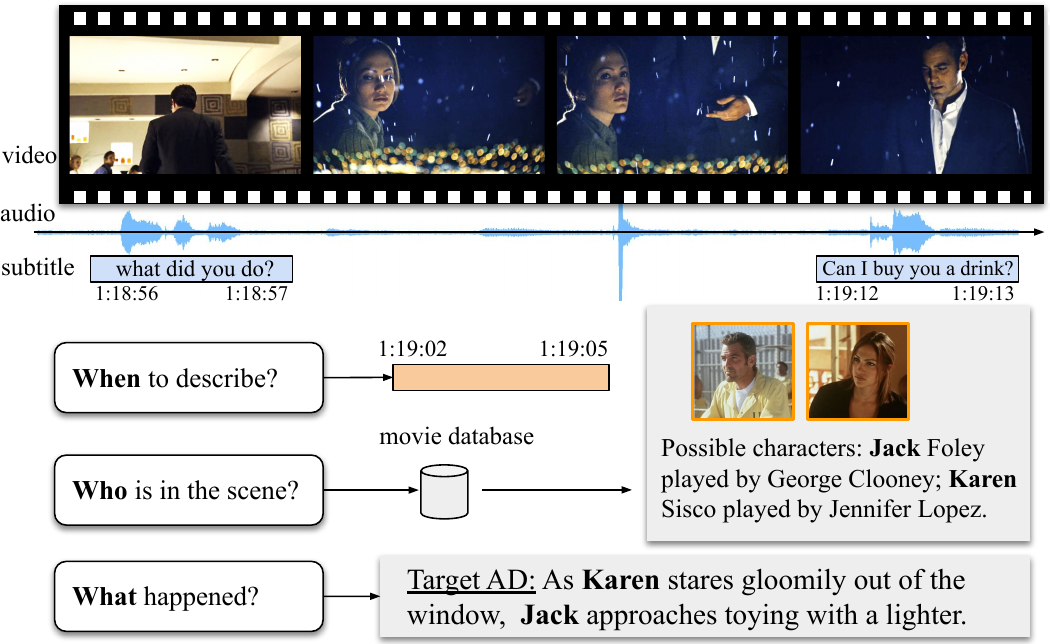}
    \caption{\textbf{AutoAD II:} We propose an automatic AD system that addresses key challenges - \textit{when} to generate AD, \textit{who} is in the scene and \textit{what} is happening visually.}
    \label{fig:teaser}
    \vspace{-4mm}
\end{figure}

Despite these benefits, the progress in generating AD is still at a very nascent stage, due to the following challenges: (a) An ideal AD generation system should perform two tasks simultaneously -- first, determine when to generate AD by proposing temporal segments; second, generate AD for the proposed segments. 
Previous works ignore the \textit{when} completely, operating on already trimmed video segments~\cite{yu2021transitional}. (b) Secondly, given the strong relevance of characters to stories~\cite{tapaswi2019,kukleva2020learning}, AD typically includes references to a character's name (\textit{who} is in the scene), their emotion, and their actions. This is particularly challenging as characters change from movie to movie. Due to anonymised test sets (LSMDC~\cite{rohrbach2015lsmdc}), the relevance of character names in AD is often ignored~\cite{yu2021transitional}. (c) Finally, AD also differs significantly from image or video captioning~\cite{lin2022swinbert, luo2020univilm, seo2022end} in that it does not need to provide descriptions of events that can be understood from the sound track alone (such as dialogue and ambient sounds) and should incorporate previous context to create a pleasurable listening experience without being repetitive or redundant. Such aspects require reasoning over multi-modal inputs (i.e., vision, text, and speech) over time while determining \textit{what} to generate. In this work we propose an AD system that focuses on all these three W's -- \textit{when}, \textit{who} and \textit{what} (Fig~\ref{fig:teaser}).

To address \textit{when}, we introduce a module to first propose temporal segments for AD. The time intervals for possible AD are constrained in that they do not overlap with the dialogue, but whether an AD is provided or not in the permissible time intervals depends on a number of factors including: the importance of the visual content to the story line, ambiguity in the audio soundtrack, and new information relative to previous AD.

For \textit{who}, we introduce an AD model that can incorporate character information {\em on-the-fly} by referring to a text-visual {\em character bank} for that movie. One of the challenges of AD is that each movie has a different set of characters (and the actors that play them) that ought to be referenced in the AD captions.
We address this by training a visual-language model to refer both to the external character bank and to the visual content of the scene when generating AD. The model can then be applied to any movie, given its cast list, without requiring retraining. This significantly improves references to characters, both in actual naming and in pronouns, in the generated AD compared to previous methods~\cite{Han23} that could only access names and pronouns present in the dialogue. Since character references appear in approximately 40\% of AD, this is an important improvement.

The final challenge is \textit{what} to generate, and involves reasoning over multimodal inputs -- images, character bank and previous AD context. We do this via a novel multimodal cross-attention architecture, which ingests proposals from the character bank, and then conditions on visual features extracted from the movie frames.

Our contributions are the following.
(1) We introduce a {\em Character Bank} to enable our AD generation model to label the characters appearing in the film.
(2) We propose a Flamingo-style~\cite{alayrac2022flamingo} architecture for the task, and compare this approach to the prompt style~\cite{mokady2021clipcap} architecture used previously for AD~\cite{Han23}.
(3) We build a model for predicting \textit{when} AD should be inserted, i.e.\ where on the timeline (using speech detection and visual cues).
(4) Given the existing challenges with captioning based metrics~\cite{fujita2020soda},  we employ a new evaluation metric for the AD content performance based on retrieval compared to other AD sentences in the movie.
(5) We significantly outperform the previous state-of-the-art on the MAD dataset \cite{soldan2022mad,Han23}.

\section{Related Work}

\noindent\textbf{Dense Video Captioning.} 
Dense video captioning is the task of temporally localising and captioning all events in an untrimmed video~\cite{krishna2017dense, wang2021end, zhou2018end}.
This differs from standard video captioning~\cite{sharma2018cc,lin2022swinbert, luo2020univilm, seo2022end}, where the goal is to produce a single caption for a given trimmed video clip. 
While most methods for dense video captioning~\cite{krishna2017dense, iashin2020better, iashin2020multi, wang2018bidirectional, wang2020event} consist of a 2-stage pipeline: a temporal localization stage followed by an event captioning stage; recent works~\cite{chadha2020iperceive, chen2021towards, deng2021sketch, li2018jointly, mun2019streamlined, rahman2019watch, shen2017weakly, shi2019dense, wang2018bidirectional, wang2021end, zhou2018end,yang2023vid2seq} jointly train the captioning and localization modules in order to improve inter-event relationships. The datasets for this task are largely obtained from web videos (\eg~YouCook2~\cite{youcook2}, ViTT~\cite{huang2020multimodal} and ActivityNet Captions~\cite{krishna2017dense}). Unlike these works, AD captions must be complementary to the audio information, tell a coherent story, and must not overlap with dialogue.

\begin{figure*}[ht!]
    \centering
    \includegraphics[width=0.95\textwidth]{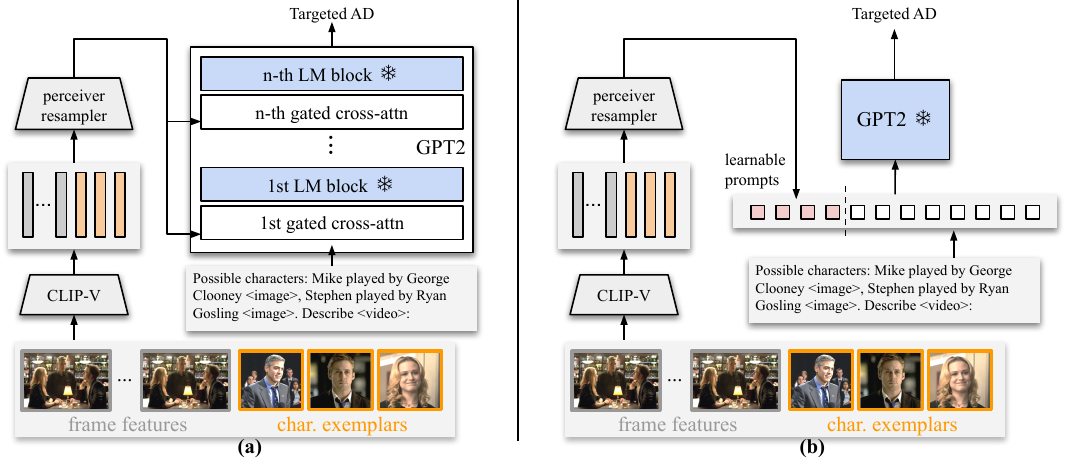}
    \vspace{-1mm}
    \caption{Architecture comparison: 
    \textbf{(a) x-attn based method}  vs. 
    \textbf{(b) prompt learning based method}.
    We use architecture \textbf{(a)} in this paper.
    The character information is fed to the model in text form. 
    Following Flamingo~\cite{alayrac2022flamingo}, we add text tags `$\langle$image$\rangle$' to indicate the association between texts and character exemplar features.
    The model can also take context AD as additional text input, by simply appending more input text tokens.
    }
    \vspace{-4mm}
    \label{fig:arch}
\end{figure*}

\noindent\textbf{Movie Understanding.} Early pioneering works exploit movies to learn actions~\cite{Laptev08}.
The LSMDC~\cite{rohrbach2015lsmdc} movie dataset sources its annotation from AD narrations and applies significant post-processing --  character name anonymization and manual timestamp refinement -- to ensure high correspondence between the short video clips and their captions. A series of short-form video tasks have since derived from LSMDC, including retrieval~\cite{Bain21}, person grounding~\cite{yu2020character}, and sequential video captioning. TPAM~\cite{yu2021transitional} tackles the latter, prompting a frozen GPT-2 with local visual features. Later works propose tasks that require more long-form modelling, including aligning movies to books~\cite{tapaswi2015book2movie,Zhu2015AligningBA} and synopses~\cite{xiong2019graph}; long video retrieval with the Condensed Movies Dataset (CMD)~\cite{bain2020cmd} and summarization~\cite{papalampidi2021movie}. \\
\noindent\textbf{Characters in Movies.}
A distinctive characteristic of movie understanding, setting it apart from other video domains, is its \textit{character-centric} nature. Thus, character recognition is a prerequisite for the task, and many works have proposed automatic identification pipelines using face, voice, and body information~\cite{Everingham06a,Tapaswi12,tapaswi2019,Nagrani17b,Brown21c}. Similar to our work,~\cite{Nagrani17b,huang2018person,brown2021automated} initialize their character recognition pipeline with actor portraits, which can be further refined with noisy image captions~\cite{huang2020caption}. Recently, CLIP has proved to be effective for zero-shot frame-level character labelling~\cite{Korbar22}, alleviating the need for complex detection pipelines, which also inspires our character identification pipeline from CLIP features. 
Dense labelling of characters in movies and TV shows enables the modeling of interactions, relationships, and intentions -- which can be formulated into classification~\cite{kukleva2020learning}, question answering~\cite{lei2018tvqa}, or captioning~\cite{lei2020tvr} tasks. Unlike these works, we use a character bank in a zero-shot practical setting for a real-world task: automatic AD generation. \\ 
\noindent\textbf{Automated Audio Description.} Visual captioning for assistive technologies is a growing area of computer vision research~\cite{bigham2010vizwiz,dognin2020image,gurari2020captioning}. Yet, generating AD for video is still a relatively unexplored area of research. Initial work~\cite{wang2021toward} applies heuristic cost-based filtering to video captioning on ActivityNet to generate diverse and relevant captions more akin to AD.
In our earlier work~\cite{Han23}, where we introduced the problem of AutoAD for movies, we provided a text-only AD corpus from over 7k movies available from the AudioVault website and used this for in-domain LLM pretraining. This resulted in substantial improvements to AD generation. We also use this dataset in this paper. However, our earlier work did not tackle the problem of \textit{when} to generate AD, assuming these segments are given a-priori, nor did it  deal with the problem of \textit{who} -- with the AD model failing to generate coherent character names, a critical component to story-coherent AD generation for long-form video content such as movies and TV shows. We address this failing in this paper.

\vspace{-1mm}
\section{New Models for Generating AD}\label{sec:method}

As in~\cite{Han23}, our method consists of adapting a large language model (LLM) for the task of generating AD. In the following sections, we describe three novel contributions: the first involves visual conditioning of multiple layers of the LLM~(Section~\ref{method:xattn}) in order to generate AD within a given time segment; the second describes a novel mechanism for incorporating character information {\em on-the-fly} that enables the model to infer a character's name in the scene~(Section~\ref{method:char}); and the third  presents a simple approach for proposing temporal segments for where in time (when) the AD should be generated (Section~\ref{method:time}).

\vspace{-1mm}
\subsection{A Visually Conditioned LM for Generating AD}\label{method:xattn}

Given a movie clip consisting of multiple
frames $\mathbf{x}_i = \{ \mathcal{I}_1, \mathcal{I}_2, ...,
\mathcal{I}_N \}$, our  aim is to produce AD text $\mathcal{T}_i$ that
describes the visual elements in a way that helps the visually
impaired follow the story line. To achieve this we build on the capabilities of a pre-trained and frozen generative 
language model (LM). Broadly, two types
of architecture are currently used to condition a LM on visual inputs: (a) by introducing additional layers into the LM that
cross-attend to the visual input (examples include Flamingo~\cite{alayrac2022flamingo}); or (b) by mapping the visual input to tokens that act as prompts for the LM (examples include ClipCap~\cite{mokady2021clipcap}). In both cases
the LM is then able to generate descriptions of the visual inputs. In our case we have multiple video frames (represented
by CLIP~\cite{clip2021} vectors) and we use a Perceiver resampler to produce a fixed sized sequence of vectors for the visual input.
The two types of architecture are illustrated in Figure~\ref{fig:arch}.

In this paper, we develop a model based on type (a), with additional cross-attention layers in the LM. We describe this
in more detail below, and demonstrate in the results that it has superior performance over type (b) in our case. We also
briefly discuss the advantages and disadvantages of the two types of architecture below.

\vspace{1mm}
\noindent\textbf{Architecture description.}  In detail, the architecture has three components: 
(i) a  CLIP encoder that generates visual features  from the input movie frames as
$\mathbf{z} = f_{\text{CLIP}}({\mathcal{I}_1,\mathcal{I}_2,...})$; (ii)
a Perceiver resampler that models the contextual
information amongst these visual features and summarizes them into
a sequence of fixed-length vectors: $\hat{\mathbf{x}} =
\mathcal{P}([\mathbf{z};\mathbf{x}])$, where $\mathbf{x}$ are learnable
latent states of the Perceiver module $\mathcal{P}$; and (iii) 
trainable cross-attention blocks that are inserted into the frozen language model.
Each cross-attention block is
controlled by a \texttt{tanh} gating mechanism, which is initialized
with zero values such that the language model maintains its original
activation at the beginning of the training as  
$\mathbf{h}_{j+1} = \mathbf{h}_j +
\texttt{tanh}\big(\texttt{XAttn}(\mathbf{h}_j,\hat{\mathbf{x}},\hat{\mathbf{x}})\big)$,
where $\mathbf{h}_j$ is the hidden vector of the $j$-th block of the
language model and $\texttt{XAttn}(q,k,v)$ denotes the cross-attention
module with its query, key and value inputs in order.  

\vspace{2mm}
\noindent \textbf{Flexibility for multimodal context.}  For our
purposes the Flamingo-like architecture offers flexibility: the input
can simply be the video frames (via the Perceiver resampler) and a
text prompt to the LM, such as `Describe $\langle$video$\rangle$:' to start the
AD generation. However, in the case that additional image and text
context is available (as in the additional character naming and image examples from the character bank, described below), then this can simply
be prepended to the prompt, and the trainable cross-attention layers learn how to correctly attend to both
the video frames and the image examples 

In contrast, for the second type of architecture where the visual input acts as a prompt to the LM, it is necessary
to train new tokens, such as BOS~\cite{Han23}, in order to separate visual prompts from text prompts and start the AD
generation.

In summary, both architectures build on frozen LMs  (previous
works~\cite{alayrac2022flamingo} show that finetuning an LLM on the task-of-interest can harm their generalization) and have trainable parameters to allow the LM to condition on the visual
input and adapt to the AD task. However, the cross-attention type of architecture offers greater flexibility and, as will be
seen, superior performance.

\subsection{Incorporating a Character Bank}
\label{method:char}


Our goal is to recognize \textit{active} characters -- defined as those appearing on-screen -- in a given
movie clip by leveraging the movie cast list from an external movie database $\mathcal{M}$, and thereby provide the information about active characters to the AD generation.
To this end,
we (i) build visual character exemplar features by
exploiting actor portrait images from $\mathcal{M}$,
further calibrated by comparing against the movie frames,
and (ii) train a character recognition module
that predicts the active characters given their exemplars
and the movie clip.

Given a long-form movie $\mathcal{V}$, the corresponding cast list can be queried from the database $\mathcal{M}$.
The character bank for this movie $\mathcal{V}$ can be written as $\mathcal{B}_{\mathcal{V}} = \{[\texttt{char}_j, \texttt{act}_j, \mathcal{A}_j]\}_{j=1}^{C}$,
where $C$ denotes the number of characters,
$\texttt{char}_j$ is the character name in the movie,
$\texttt{act}_j$ is the actor name, and 
$\mathcal{A}_j$ is the actor's portrait image from the movie database.
Below are two example items in a character bank:
\vspace{-2mm}
\begin{align*}
\{
&[\text{Jack Dawson}, \text{Leonardo DiCaprio}, \mathcal{A}_{\text{LD}}],\\
&[\text{Rose DeWitt-Bukater}, \text{Kate Winslet}, \mathcal{A}_{\text{KW}}],...
\}
\vspace{-2mm}
\end{align*}

\begin{figure}[t!]
    \centering
    \includegraphics[width=0.47\textwidth]{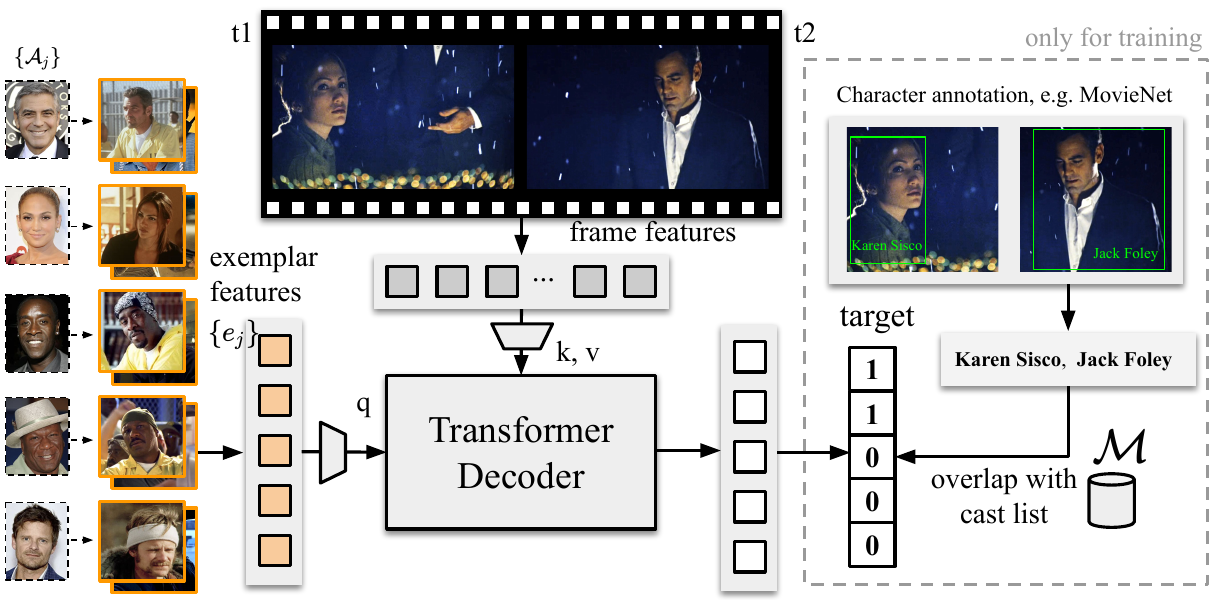}
    \caption{
    \textbf{Character recognition module}:
    Given character exemplar features for $C$ characters $\{e_j\}_{j=1}^{C}$,
    and movie frame features for a given clip,
    we formulate a binary classification problem
    to determine whether each character is active in the scene or not. 
    From left to right: portrait images determine a within-movie exemplar for each character, and each of these exemplars provides a query to the transformer. The frame features provide the keys and values for the transformer.
    The output of each query is used to determine the binary question of whether that character is in the clip or not. 
    The module is trained using MovieNet data where characters are annotated for the frames of a clip. 
    The binary labels are formed by checking the MovieNet character annotations against the cast list in our movie database.
    }
    \label{fig:char_arch}
    \vspace{-1mm}
\end{figure}

\noindent\textbf{Calibrating the actor portrait feature.}
An actor's portrait image can differ considerably in appearance from the character in the movie due to various factors, such as hairstyle, makeup, dress, ageing, or camera viewpoint~\cite{Nagrani17b}.
In particular, for older movies with different dressing styles and fewer close-up shots, actor portraits might lie very far from the movie's frame in the feature space.
To overcome this issue, we propose a calibration step.
Instead of using the image features from the actor's portrait,
we retrieve the top-$k$ nearest frames within the same movie,
and average the frame features to create an exemplar for that
character.
Specifically, let $\mathbf{z}_{\mathcal{V}} = f_{\text{CLIP}}(\mathcal{V})$ denote the sequence of visual features of the movie $\mathcal{V}$, 
and given a portrait image of actor $j$ as $\mathcal{A}_j$,
we first compute its visual feature $z_{j} = f_{\text{CLIP}}(\mathcal{A}_j)$, and compare it against
$\mathbf{z}_{\mathcal{V}}$ via cosine similarity.
The character exemplar feature of actor $j$ in the movie $\mathcal{V}$
can then be computed by:
\vspace{-0.5mm}
\begin{equation*}
    e_j = \frac{1}{k} \sum_{}{
    \mathbf{z}_{\mathcal{V}}\Bigg[\text{top-}k\big(
    \frac{z_{j}^{\top}\mathbf{z}_{\mathcal{V}}}
    {|z_{j}| \cdot |\mathbf{z}_{\mathcal{V}}|}
    \big)\Bigg],
    }
\vspace{-1.0mm}
\end{equation*}

\noindent where top-$k$ finds the indices of the $k$ most similar frames
and $[.]$ symbol means the indexing operation.
In the~\app,
we show this calibration procedure
(i.e., replacing $z_j$ with $e_j$)
is essential for constructing reliable character banks. \\

\begin{figure*}[ht!]
  \centering
  \subfigure[]{\includegraphics[scale=0.43]{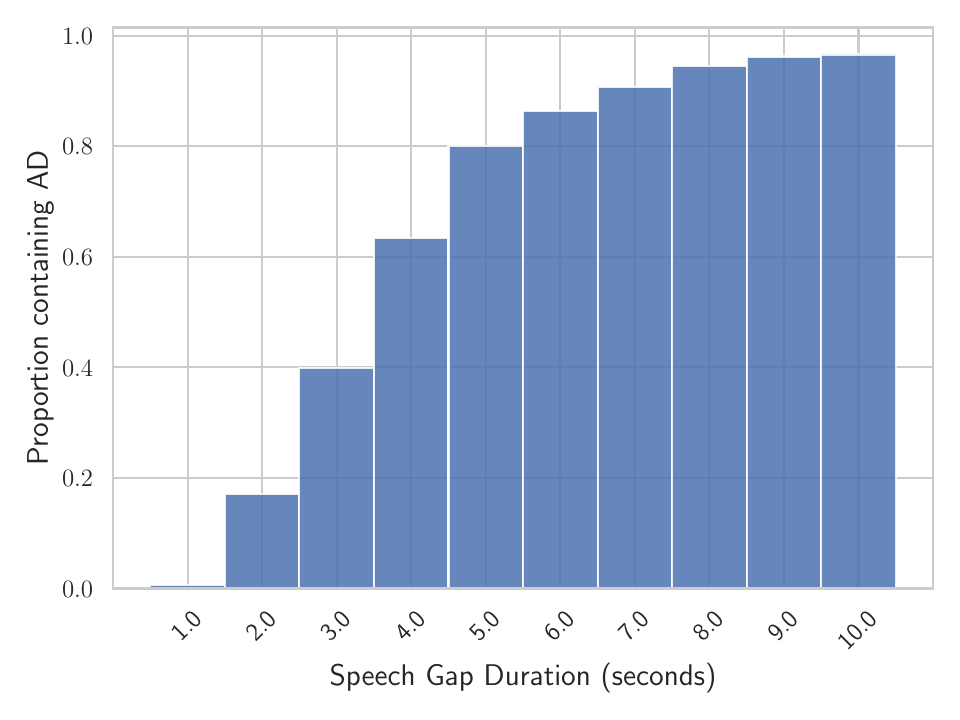}\label{fig:ad_hist}}\hspace{3em}
  \subfigure[]{\includegraphics[scale=0.56]{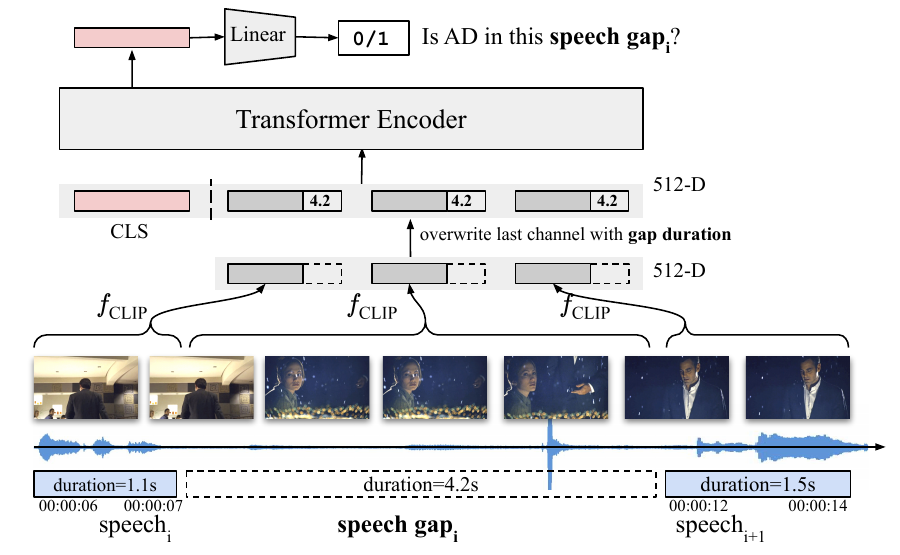}}
  \vspace{-1mm}
  \caption{\textbf{(a) Proportion of speech gaps containing AD relative to their duration} -- very short speech gaps rarely contain AD, and large speech gaps nearly always contain AD. The statistics are from the MAD training set. \textbf{(b) Architecture for AD temporal proposal classification.} Given a speech gap, the model classifies whether or not AD should be inserted in the gap, taking visual and duration cues as input.}\label{fig:AD_hist}
  \vspace{-3mm}
\end{figure*}

\noindent\textbf{Recognizing characters in the movie clip.}
Not all characters appear on-screen at the same time.
With a character bank $\mathcal{B}_{\mathcal{V}}$ for the movie $\mathcal{V}$, our goal is to recognize the \emph{active} characters that appear between times $t_1$ and $t_2$ to enable naming them in the AD generation.
This character recognition task may be achieved by face detectors~\cite{deng2020retinaface},
or even speaker recognition from voice \cite{Xie19a}.
However, for the movie datasets used in this work, 
the absence of raw frames prohibits the use of face detection,
and the characters mentioned in AD may not necessarily be speaking. Instead, we propose to use a character recognition module based purely on frame-level visual features and the character bank information $\mathcal{B}_{\mathcal{V}}$.

As shown in Fig.~\ref{fig:char_arch},
both the exemplar features for each character $\{e_j\}_{j=1}^{C}$ and the movie frame features $\mathbf{z}_{\mathcal{V}[t_1,t_2]}$ are first fed to a linear projection layer, which aims to project general visual features onto a face feature space~\cite{Korbar22}.
Then a relatively shallow (2-block) transformer decoder 
takes both projected features
and outputs a probability for each character on
whether they appear between times $[t_1,t_2]$.
This module is trained with a binary classification loss.
The labels for training are obtained from face annotations available in the MovieNet~\cite{huang2020movienet_short} dataset.
In Sec.~\ref{exp:char_recog},
we compare our model with a baseline of simply thresholding the similarity between the character's exemplar and movie frame features with a scalar $\alpha$. The experiment shows that the  transformer decoder module outperforms this baseline. In the~\app, we also experiment with training from labels obtained by running named entity recognition (NER)~\cite{nadeau2007survey} on the AD annotation, and show that this performs worse than using labels obtained from MovieNet.

\begin{figure}[h!]
    \centering
    \includegraphics[width=0.45\textwidth]{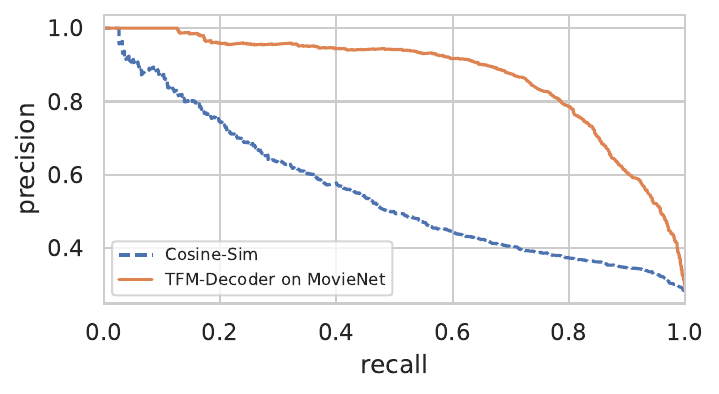}
    \vspace{-1mm}
    \caption{
    The Precision-Recall curve for the character recognition methods, 
    computed on 4 MAD-eval movies that have character annotations from MovieNet.
    {We compare two methods: thresholding actor-movie cosine similarity,
    and learned transformer decoder on MovieNet.
    The precision/recall is calculated on a per-character basis,
    \ie~the precision/recall of the cosine thresholds to correctly find a character name mentioned in the AD. More baselines are described in the~\app.}}\vspace{-2mm}
    \label{fig:pr_curve}
\end{figure}

\noindent\textbf{Other approaches.}
In our earlier work~\cite{Han23}, we mined character names from subtitles using NER and provided these as prompts to the AD generation model -- but this failed to reference character names effectively.
The failure may be because character names occur sparsely in subtitles (Based on the MAD-train movies, approximately 13\% of subtitle sentences contain character names, compared to 41\% of AD, see~\app~Table~\ref{table:appendix:freq_ad} and Table~\ref{table:appendix:freq_sub}) or because the names found may refer to off-screen characters.
In Figure~\ref{fig:pr_curve}, 
we show that the movie cast list (when narrowed down to those that appear in the scene with our character recognition module) provides high precision and recall of active on-screen characters.

\noindent\textbf{Using the character bank for AD generation.}
A trained character recognition module can recognize the \emph{active} characters in any video clip $\mathcal{V}_{[t_1,t_2]}$.
Next, we feed this character information into our AD generation pipeline.

In Sec.~\ref{method:xattn},
we introduce a versatile cross-attention-based architecture which supports textual and other multi-modal inputs.
We feed in character information to the model mainly by \emph{text prompting}.
In more detail, given a character list for the movie clip $\mathcal{V}_{[t_1,t_2]}$, we explore three different ways of supplying the active characters in the scene. Let's assume
[$\texttt{char}_1$, $\texttt{char}_2$] are recognized as active. The prompting templates are then:
\vspace{-1mm}
\begin{enumerate}[itemsep=-0.2em]
\item {``possible characters: $\texttt{char}_1$, $\texttt{char}_2$.''}
\item {``possible characters: $\texttt{char}_1$ played by $\texttt{act}_1$; 
    $\texttt{char}_2$ played by $\texttt{act}_2$.''}
\item {``possible characters: $\texttt{char}_1$ played by $\texttt{act}_1$ $\langle$image$\rangle$; $\texttt{char}_2$ played by $\texttt{act}_2$ $\langle$image$\rangle$.''}
\end{enumerate}
\vspace{-1mm}

Note that in method (3), the $\langle$image$\rangle$ tag is purely in the text form; therefore, in this setting, we feed in the character 
exemplar features $[e_1, e_2]$ in the corresponding order to the perceiver resampler, such that it can learn the association between the character's identity and the movie clip.

\subsection{Proposing AD Temporal Segments}
\label{method:time}

An ideal AD system must not only generate high quality AD narrations (\textit{what}), but must also decide \textit{when} to generate AD. The Web Content Accessibility Guidelines 2.0~\cite{caldwell2008web} outlines specific criteria for successful AD: (i) it must only be added during existing pauses in dialogue; and (ii) it need not be added when all of the video information is already provided in existing audio.

In practice, long pauses in dialogue and the subjectivity of the second guideline mean these provide rather weak constraints on the timing of AD, resulting in large variations between human-generated AD timestamps for the same movie\footnote{An analysis on Audio Descriptions and inter-annotator agreement is provided in the~\app.\label{foot:ad_analysis}}.
Such variation makes it difficult to learn and evaluate fine-grained model predictions of proposed AD temporal segments. Therefore, we formulate the temporal proposal task into one of binary classification: \textit{given an existing pause in dialogue, should AD be inserted in the pause?} This coarse-grained formulation has much higher inter-annotator agreement and inherently satisfies the first guideline for generating AD$^{\ref{foot:ad_analysis}}$.

Given a long-form movie $\mathcal{V}$, our goal is to identify inactive speech regions and classify whether or not they should contain audio description (AD). Thirty second intervals are extracted from the movie. Automatic speech recognition (ASR) is applied to the audio stream $A$ of each interval to obtain a set of speech segments $S = {s_1,...,s_N}$, where each $s_i = (t_i^0, t_i^1)$ indicates the start and end times of a speech utterance. The text from the ASR segments $s_i$ are given as input to a BERT encoder, prefixed and suffixed with discrete timestamp tokens $\tau \in {\texttt{<|t00|>},...,\texttt{<|t60|>}}$ denoting their start and end times rounded to 0.5 seconds. The gaps between utterances are represented by inserting $\texttt{<|mask|>}$ tokens between timestamped segments.
 In addition to speech, we also provide visual features from CLIP~\cite{radford2022whisper}, sampled every second from the context window, and append these to the input text sequence. 

A binary classification head is then applied to each $\texttt{<|mask|>}$ token to predict whether the gap should contain AD. The model is trained end-to-end using binary cross-entropy loss. At inference time, this model is applied in a sliding window fashion to the full movie $\mathcal{V}$. An overview of the method is provided in Figure~\ref{fig:ad_hist}. Further details are provided in Appendix~\ref{app:temp_class}.

By analysing the distribution of AD data (Figure~\ref{fig:ad_hist}), we find that whether or not AD is contained within a given speech gap is highly correlated with the duration of said gap. In fact, gaps of two seconds or less contain AD only 17\% of the time. At the other extreme, gaps of 6 seconds or more contain AD 85\% of the time. Due to such strong duration correlations, we restrict the prediction task to speech gaps between two and five seconds. The classification of whether to insert AD within shorter or longer speech gaps can be obtained via a hard-coded rule. The effect of timestamp tokens and visual features are given in Section~\ref{exp:time}.

\section{Implementation Details}
\subsection{Training Data}
\noindent\textbf{MAD~\cite{soldan2022mad}} is a movie audio description dataset consisting of movie frame features and timed AD in the text form. We follow~\cite{Han23} and use 488 movies as the training set. Specifically for AD, we use the same preprocessing pipeline proposed in~\cite{Han23} to obtain high-quality ASR outputs. We use the `named' version of MAD dataset.
\textbf{AudioVault-AD~\cite{Han23}} is a text-only corpus of AD for 7057 movies downloaded from the AudioVault website. The movies are not included in MAD dataset. 
We use the AudioVault-AD for text-only pretraining.
\textbf{WebVid~\cite{Bain21}} is a dataset of 2.5M captioned short videos for visual-only pretraining.
We find the NER from both LSMDC-train and MAD-train contain non-trivial noise, despite the one for LSMDC-train having been manually verified. 
\textbf{MovieNet}~\cite{huang2020movienet_short}
is a movie dataset providing movie keyframes and various annotations including character names for each keyframe.
We choose an overlap of MovieNet movies with MAD training movies to train the character recognition module.

\subsection{Testing Data}
\noindent\textbf{MAD-eval~\cite{Han23}} consists of 10 movies for evaluating AD captioning from the LSMDC validation and testing set. The timestamps from LSMDC are manually edited to ensure high visual correspondence with the caption. We treat this as our standard evaluation for measuring AD caption quality.

\noindent\textbf{MAD-t-eval} is our proposed benchmark for evaluating AD time point prediction. The edited timestamps in \textit{MAD-eval} are not appropriate for measuring temporal proposals because they are expanded and often overlap with speech segments. Therefore we evaluate time prediction models on \textit{MAD-t-eval}, consisting of three movies (from MAD-eval) where the AD and their original timestamps are sourced from Audiovault and manually verified. We restrict the evaluation to speech gaps with a duration between two and five seconds, resulting in 530 gaps across the three movies.

\subsection{Collecting Character Banks}
The character information for movies can be collected from online databases
or review websites like IMDb\footnote{https://www.imdb.com/}.
In detail, for each movie in Audiovault, MAD-train and the MAD-eval datasets,
we download the top 10 cast information from IMDb including the actor names, their character role name, and the actor portrait image. 
Full details are provided in~\app.

\subsection{Training \& Inference Recipe}

In this section, we first outline the architectures used for each module in the AD captioning system; we then describe how each module individually is pretrained; and finally we describe the finetuning and inference details for the full AD captioning system.

\subsubsection{Architectural {components}}

We give a summary here with fuller details in the Appendix.

\noindent\textbf{AD generation model}~(Section~\ref{method:xattn})
is built on top of \emph{GPT2-small}, specifically the open-source version from HuggingFace. We insert an X-attn block after \emph{each} of the transformer block of GPT-2. 
The perceiver resampler has two transformer decoder blocks with 10 latent vectors. 
For the visual encoder, we use CLIP~\cite{clip2021} ViT-B/32 model which extracts $512$-d features for each movie frame. These features are provided by the MAD dataset \cite{soldan2022mad}.

\noindent\textbf{Character recognition module}~(Section~\ref{method:char}) consists of a linear layer and a 2-block transformer decoder. It takes the movie character exemplar features $\{e_j\}$ and movie clip features as input, and outputs a probability for each exemplar feature.

\noindent\textbf{AD temporal proposals}~(Section~\ref{method:time}). For VAD, we use the \textit{pyannote} model~\cite{bredin2020pyannote}. For the temporal proposal classification model, we use a 3-layer transformer encoder with sin-cos positional embeddings. For the visual features we use CLIP ViT-B/32~\cite{clip2021}.

\subsubsection{Pretraining recipe}
To overcome the limited amount of paired AD training data, we follow~\cite{Han23} and perform partial data pretraining for each component in our modular architecture.

\noindent\textbf{GPT-2}~(Section~\ref{method:xattn}). We follow~\cite{Han23} and perform secondary in-domain pretraining of GPT-2 on the Audiovault text-only corpus to match the text distribution for AD generation.

\noindent\textbf{Video captioning}~(Section~\ref{method:xattn}). We pretrain the cross-attention visual captioning blocks on 2.5M video-text pairs from WebVid~\cite{Bain21}, while keeping the GPT-2 LM block weights frozen.

\noindent\textbf{Character recognition module}~(Section~\ref{method:char}). The module is trained on character name labels from MovieNet~\cite{huang2020movienet_short}.

\subsubsection{Finetuning \& Inference}

\noindent\textbf{AD captioning}~(Section~\ref{method:xattn}). 
With the recognized active character list as an additional input
and model parameters partially pretrained,
the AD generation model is finetuned on MAD-train
with an AdamW~\cite{loshchilov2017adamw} optimizer and $10^{-4}$ learning rate.
For output text sampling, we use beam search with the beam size of 5 and report results by the top-1 beam-searched outputs, since it performs slightly better than greedy search on multiple scenarios.
The full training details are in~\app.

\noindent\textbf{AD temporal proposals}~(Section~\ref{method:time}). The pretrained BERT (base-uncased) is finetuned on the MAD dataset for three epochs, with a BCE loss and an AdamW~\cite{loshchilov2017adamw} optimizer of learning rate $10^{-4}$. The classification task at training and inference is restricted to speech gaps with durations between 2-6 seconds.

\section{Experiments}
The experimental section is organised as follows:
we start by describing the evaluation metrics in Section~\ref{exp:evaluation};
then in Section~\ref{exp:result}, we demonstrate the effectiveness of our proposed architecture and training strategy, based on the groundtruth AD time segments, for example, 
visually conditioned LM, effect of character bank, and partial-data pretraining;
in Section~\ref{exp:time}, we evaluate on the temporal proposal, 
and present qualitative results in Section~\ref{exp:qualitative}.

\subsection{Evaluation Metrics}
\label{exp:evaluation}

\noindent\textbf{Classic metrics for text generation.} 
We adopt classic captioning metrics to compare the generated AD to the ground-truth AD,
namely, ROUGE-L~\cite{lin2004rouge}~(\textbf{R-L}) and CIDEr~\cite{vedantam2015cider}~(\textbf{C}). 

\vspace{3pt}
\noindent\textbf{Retrieval-based metric for text sequence generation.}
We propose a new recall-based metric: `Recall@$k$ within $N$ neighbours' (\textbf{R@k/N}). In detail, given two sequences of generated texts and ground-truth (GT) texts in their temporal order, for each generated text at time point $[t_1,t_2]$,  we compute the Recall@$k$ with $N$ adjacent GT texts, then average the score. 
To compute recall, we use the BertScore~\cite{zhang2019bertscore} as the text similarity measure.
There are two benefits of this metric:
(i) Classic captioning metrics like CIDEr or ROUGE-L are mainly based on n-gram accuracy, which tends to over-penalise the system on linguistic text variations, 
\ie because there are multiple ways to express the same meaning. 
The retrieval-based \textbf{R@k/N} metric is less affected by these low-level variations in the text.
{(ii)
the metric is operated within a window of $N$ neighbouring texts along the time axis, which considers the arbitrary text positioning of long sequence captioning.
}

\vspace{3pt}
\noindent\textbf{Metrics for character recognition and time proposal.}
The character recognition (described in Section~\ref{method:char}) 
and the time segment proposal tasks (described in Section~\ref{method:time}) are formulated as multi-label and binary classification problems respectively
We report ROC-AUC and Average Precision for the classifiers, with class macro-averaging for the multi-label case.

\subsection{Audio Description on GT segments}\label{exp:result}

This section focuses on the effectiveness of each proposed component in the AD generation pipeline, based on the \emph{ground-truth} AD time segments, as shown in Table~\ref{tab:char_bank_abl}.

\vspace{-8pt}
\subsubsection{Architecture comparison}
\vspace{-3pt}
We investigate two ways for conditioning a pre-trained and frozen generative language model (LM) with visual inputs, that is, 
(a) by introducing additional layers into the LM that cross-attend to the visual input, 
or (b) by mapping the visual input to tokens that
act as prompts for the LM.
Comparing rows `B1 vs A1' and `B4 vs A2' in Table~\ref{tab:char_bank_abl}, 
the architecture with newly introduced cross-attention outperforms the prompting-based architecture 
both with or without the character bank inputs. The performance gain comes from greater interaction between visual and textual features by its interleaved design.

\begin{table}[t]
\centering
\footnotesize
\vspace{2mm}
\setlength{\tabcolsep}{1pt}
\begin{tabular}{@{}ccclllllll@{}}
\toprule
\multirow{2}{*}{Exp.} & \multirow{2}{*}{\begin{tabular}[x]{@{}c@{}}AD\\Context\end{tabular}} & \multirow{2}{*}{PT} & \multirow{2}{*}{Arch} & \multicolumn{4}{c}{CharBank Settings} & \multirow{2}{*}{R-L} & \multirow{2}{*}{C} \\
 & & &  & Source & Char. & Act. & Exem. &  &  \\ \midrule
A1 & \xmark & \xmark & Prompt & -         & \xmark & \xmark & \xmark & 9.3  & 6.7 \\
A2 & \xmark & \xmark & Prompt & recog.    & \cmark & \cmark & \cmark & 10.4 & 11.0 \\ \midrule
B1 & \xmark & \xmark & X-Attn & -         & \xmark & \xmark & \xmark & 9.7  & 10.0 \\
B2 & \xmark & \xmark & X-Attn & recog.    & \cmark & \xmark & \xmark & 10.8 & 14.2 \\
B3 & \xmark & \xmark & X-Attn & recog.    & \cmark & \cmark & \xmark & 11.1 & 15.0 \\
B4 & \xmark & \xmark & X-Attn & recog.    & \cmark & \cmark & \cmark & 12.7 & 18.3 \\
B5 & \xmark & \xmark & X-Attn & full-cast & \cmark & \cmark & \cmark & 10.9 & 14.9 \\
\midrule
C1 & \xmark & AV\&WV  & X-Attn & recog.    & \cmark & \cmark & \cmark & 13.1 & 19.2 \\
C2 & \cmark (recurrent) & AV\&WV  & X-Attn & recog.    & \cmark & \cmark & \cmark & 13.4 & 19.5 \\
\bottomrule
\end{tabular}
\vspace{0.5mm}
\caption{\textbf{Ablations for AD generation.} 
We ablate the effect of the cross-attention module and character bank, 
and show the effect of partial-data pretraining.
All models are trained on MAD-train-\texttt{named} and evaluated on MAD-eval-\texttt{named}.
Performance is reported in terms of ROUGE-L~(R-L) and CIDEr~(C).}
\label{tab:char_bank_abl}
\vspace{-0.2cm}
\end{table}
\begin{table}[h]
    \small
    \centering
    \begin{tabular}{ccc}
    \toprule
    Methods                          & ROC AUC       & Average Precision \\ \midrule
    Cosine-Sim                      & {0.72}          &  {0.55}             \\
    TFM Decoder                     & {\textbf{0.93}} & {\textbf{0.87}}     \\
    \bottomrule
    \end{tabular}
    \vspace{1mm}
    \caption{{We compare different methods for recognising characters in a clip, reported on 4 MAD-eval movies that have character annotations from MovieNet.}}
    \label{table:char_recog}
    \vspace{-1mm}
\end{table}

\vspace{-8pt}
\subsubsection{Effect of character bank}\label{exp:char_recog}
\vspace{-4pt}
Here, we start by investigating the effect of incorporating the character bank in three different ways as discussed in Section~\ref{method:char}, followed by comparing our proposed character recognition module with a na\"ive baseline.
\begin{figure*}[t]
    \centering
    \includegraphics[width=1.0\textwidth]{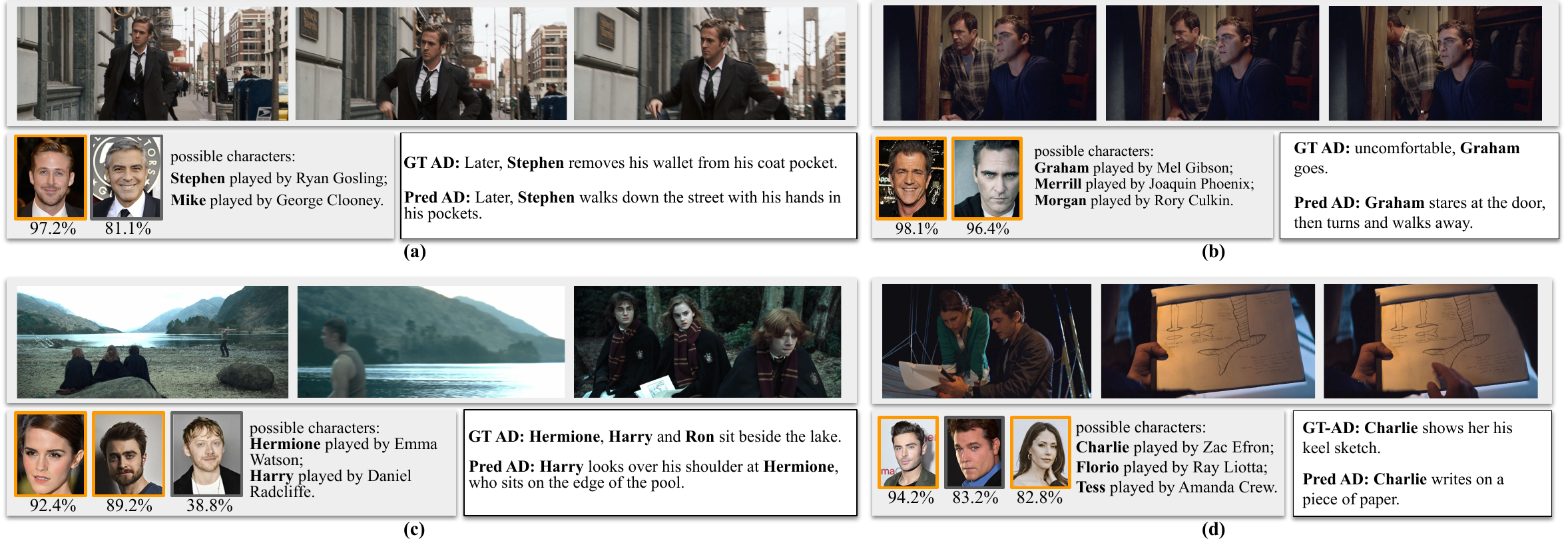}
    \caption{\small{Qualitative results of our method.
    For a given movie clip, the character recognition module can recognize active characters on-the-fly and its results are fed into the AD generation pipeline.
    Note that for character recognition, we simply threshold the `active' characters in the scene with a probability of 50\%. The probability shown below characters' portraits is the output of our recognition module, with correctly recognized characters marked using an {\color{orange} orange border}.
    For visualization purpose, we show the character's IMDb portrait image but the model actually takes in exemplar features as input.
    Three frames are shown for each movie clip.
    To illustrate the effect of character bank, this model is trained \emph{without} context AD -- that is, naming information is only from the character bank.
    The movies are from: \textbf{(a)}: Ides of March (2011), 
    \textbf{(b)}: Signs (2002),
    \textbf{(c)}: Harry Potter and the Goblet of Fire (2005),
    \textbf{(d)}: Charlie St.\ Cloud (2010).}}
    \label{fig:qualitative}
    \vspace{-1mm}
\end{figure*}

\vspace{3pt}
\noindent \textbf{Choices for exploiting character bank.}
By default, the model takes the predicted `active' characters in the scene from the character recognition module.
From the comparison rows `B1-4', we can draw the following observations:
(i) injecting character names gives a clear performance gain~(`B2 vs B1'),
highlighting the dependency of the AD task on character names;
(ii) inputting additional actor names only brings marginal improvements~(`B3 vs B2'),
we conjecture this is because CLIP have seen large number of celebrities' picture-names pairs at pre-training stage, the visual features have thus already encoded such information;
(iii) feeding in characters' exemplar features improves the performance (`B4 vs B3'), 
showing the complementary nature of visual-textual features;
(iv) presenting the full cast list (\ie all 10 characters downloaded from IMDb) as character bank 
leads to inferior performance, as shown by comparison between B4 and B5. 
This is because the full cast list introduces irrelevant characters to the AD generation pipeline and harms the training, especially with a limited size of training data. 
This comparison also shows the necessity and effectiveness of our character recogition module, 
which provides a higher-quality character list to aid AD generation.

\vspace{3pt}
\noindent \textbf{Character recognition module.}
We compare the proposed character recognition module (described in Fig.~\ref{fig:char_arch}) with the baseline method that simply thresholds CLIP similarities between exemplar features and frame features. 
Specifically, we train the character recognition module on 550 MovieNet movies with character annotations,
and report results on 4 MAD-eval movies that have character annotations from MovieNet.
Since the task of recognizing active characters is a binary classification,
we report ROC-AUC and Average Precision, as shown in Table~\ref{table:char_recog},
our proposed character recognition module clearly outperforms the baseline by a large margin.
More details and discussion on a PR curve are provided in the Appendix. 



\begin{table}[!t]
\centering
\small
 \setlength{\tabcolsep}{6pt}
 \vspace{2mm}
\begin{tabular}{@{}lcc@{}}
\toprule
\multicolumn{1}{c}{Methods} & \multicolumn{1}{c}{ROC AUC} & \multicolumn{1}{c}{Average Precision} \\ \midrule
Baseline (Duration) & 0.70 & 0.53 \\
TFM (Visual) & 0.71 & 0.52 \\
TFM (Visual+Duration) & \textbf{0.78} & \textbf{0.61} \\
\bottomrule
\end{tabular}
\vspace{0.5mm}
\caption{Results on the binary AD temporal proposal task on the \texttt{MAD-t-eval} benchmark where TFM refers to the transformer encoder architecture.
}
\label{tab:time_proposal}
\vspace{-2mm}
\end{table}
\subsubsection{Partial-data pretraining and context}
Following~\cite{Han23}, we also pre-train our model with partial-data, 
for example, AudioVault and WebVid, as well as incorporate previous AD as context for the model.
The results show that AD generation can be further improved by combining these methods,
showing that our newly introduced cross-attention module and character bank are orthogonal to the contributions in~\cite{Han23}.

\subsection{Temporal proposal results}\label{exp:time}
{
In Table~\ref{tab:time_proposal} we compare the different time proposal classification models with the baseline threshold method.
We see that training a transformer encoder architecture with both visual and speech gap duration information as inputs brings substantial improvements (0.53 AP to 0.61 AP)
}

\subsection{Qualitative Results}\label{exp:qualitative}
Fig.~\ref{fig:qualitative} shows four qualitative examples.
It shows that the character recognition module is able to recognize active characters reasonably well, and the AD generation module can associate the active characters with the descriptions.
Note that even if the character recognition module proposes incorrect characters,
the AD generation pipeline has learned to \emph{ignore} such irrelevant characters for AD generation, 
such as `Mike' in sample \textbf{(a)} and Morgan in sample \textbf{(b)}.
Given that a large portion of AD sentences (41\%) contains human identity like those samples,
recognizing characters is an essential capability for high-quality AD generation. 
More examples in the~\app.

\subsection{Comparison with state-of-the-art}\label{exp:sota}
In Table~\ref{table:sota-mad}, we report AD captioning results on the MAD-eval benchmark and achieve state-of-the-art performance by considerable margins across both the local and recurrent settings.
Note that our method \emph{without} context AD or AV/WebVid pretraining already surpasses 
AutoAD-I (CIDEr 18.3 vs 14.3). Adding partial data pretraining on AV/WebVid and context AD further increases the performance (CIDEr 19.5 vs 14.3).
Note, one issue that affects the evaluation is that although we might correctly identify a character, and they are referred to in the AD, the actual name may differ since
 a character may be named in a multitude of ways,~\eg 
first name only -- `Albus', or last name with a prefix -- `Mr.\ Dumbledore', 
their professions, titles or pronouns -- `Professor', `Prof. Dumbledore' or `He', 
their relationships to other characters -- `Aberforth's brother', or other nicknames etc. We leave the resolution of this issue to future work.

\begin{table}[t]
\footnotesize
    \centering
    \setlength{\tabcolsep}{3pt}
    \vspace{1mm}
    \begin{tabular}{lll|ccl}
    \toprule
    Methods & {Time window} & Pretrain Data                             & R-L  & C   & \multicolumn{1}{l}{R@5/16}\\ \midrule
    ClipCap~\cite{mokady2021clipcap} & local & CC3M                                      & 8.5  & 4.4  & 36.5$^{*}$  \\
    AutoAD-I~\cite{Han23} & local    & WebVid           & 9.9  & 10.0 & 38.2$^{*}$ \\
    AutoAD-I~\cite{Han23} & local    & AV \& WebVid              & 10.3  & 12.1 & 39.8$^{*}$ \\
    \textbf{Ours} & local                     & None             & 12.7 & 18.3 & 45.6 \\
    \textbf{Ours} & local                     & AV \& WebVid     & \textbf{13.1} & \textbf{19.2} & \textbf{51.3} \\
    \midrule
    AutoAD-I~\cite{Han23} &recurrent & AV \& WebVid & 11.9      & 14.3      & 42.1$^{*}$ \\
    \textbf{Ours}   &recurrent               & AV \& WebVid    & \textbf{13.4} & \textbf{19.5}    & \textbf{50.8}   \\
    \bottomrule
    \end{tabular}
    \vspace{1mm}
    \caption{Comparison with other methods on \texttt{MAD-eval} benchmark under both the local (without AD context) and recurrent (with previously predicted AD as context) settings. $^{*}$Denotes results re-implemented by us using the same evaluation setting.}
    \vspace{-1mm}
    \label{table:sota-mad}
\end{table}

\vspace{8mm}
\section{Discussion and Future Work}
Taken together this paper has proposed
all the elements needed for a fully automated AD system: when to produce AD, what it should contain, and who it should describe (naming). 
Note these sub-tasks can probably be done jointly by using a transformer decoder with special time tokens, such as Whisper~\cite{radford2022whisper} or Vid2Seq~\cite{yang2023vid2seq}. 
Predicting accurate timestamps for such architectures~\cite{bain2022whisperx},
modelling long-term dependency and leveraging multi-modal information
are exciting challenges towards human-level movie understanding.

\vspace{3mm}

\noindent\textbf{Acknowledgements.}
We thank the AudioVault team for their priceless contribution of Audio Description archives.
This research is funded by EPSRC PG VisualAI EP/T028572/1, 
and ANR-21-CE23-0003-01 CorVis.

\clearpage
{\small
\bibliographystyle{ieee_fullname}
\bibliography{bib/shortstrings,bib/vgg_local,bib/vgg_other}

\begin{thebibliography}{10}\itemsep=-0.4pt

\bibitem{alayrac2022flamingo}
Jean-Baptiste Alayrac, Jeff Donahue, Pauline Luc, Antoine Miech, Iain Barr,
  Yana Hasson, Karel Lenc, Arthur Mensch, Katie Millican, Malcolm Reynolds,
  et~al.
\newblock Flamingo: a visual language model for few-shot learning.
\newblock In {\em NeurIPS}, 2022.

\bibitem{bain2022whisperx}
Max Bain, Jaesung Huh, Tengda Han, and Andrew Zisserman.
\newblock {WhisperX}: Time-accurate speech transcription of long-form audio.
\newblock In {\em Interspeech}, 2023.

\bibitem{bain2020cmd}
Max Bain, Arsha Nagrani, Andrew Brown, and Andrew Zisserman.
\newblock Condensed movies: Story based retrieval with contextual embeddings.
\newblock In {\em Proc. ACCV}, 2020.

\bibitem{Bain21}
Max Bain, Arsha Nagrani, G{\"u}l Varol, and Andrew Zisserman.
\newblock Frozen in time: A joint video and image encoder for end-to-end
  retrieval.
\newblock In {\em Proc. ICCV}, 2021.

\bibitem{bigham2010vizwiz}
Jeffrey~P Bigham, Chandrika Jayant, Hanjie Ji, Greg Little, Andrew Miller,
  Robert~C Miller, Robin Miller, Aubrey Tatarowicz, Brandyn White, Samual
  White, et~al.
\newblock Vizwiz: nearly real-time answers to visual questions.
\newblock In {\em Proceedings of the 23rd annual ACM symposium on User
  interface software and technology}, pages 333--342, 2010.

\bibitem{bredin2020pyannote}
Herv{\'e} Bredin, Ruiqing Yin, Juan~Manuel Coria, Gregory Gelly, Pavel
  Korshunov, Marvin Lavechin, Diego Fustes, Hadrien Titeux, Wassim Bouaziz, and
  Marie-Philippe Gill.
\newblock Pyannote. audio: neural building blocks for speaker diarization.
\newblock In {\em Proc. ICASSP}, pages 7124--7128. IEEE, 2020.

\bibitem{brown2021automated}
Andrew Brown, Ernesto Coto, and Andrew Zisserman.
\newblock Automated video labelling: Identifying faces by corroborative
  evidence.
\newblock In {\em 2021 IEEE 4th International Conference on Multimedia
  Information Processing and Retrieval (MIPR)}, pages 77--83. IEEE, 2021.

\bibitem{Brown21c}
Andrew Brown, Vicky Kalogeiton, and Andrew Zisserman.
\newblock Face, body, voice: Video person-clustering with multiple modalities.
\newblock In {\em ICCV 2021 Workshop on AI for Creative Video Editing and
  Understanding}, 2021.

\bibitem{caldwell2008web}
Ben Caldwell, Michael Cooper, Loretta~Guarino Reid, Gregg Vanderheiden, Wendy
  Chisholm, John Slatin, and Jason White.
\newblock Web content accessibility guidelines (wcag) 2.0.
\newblock {\em WWW Consortium (W3C)}, 290:1--34, 2008.

\bibitem{chadha2020iperceive}
Aman Chadha, Gurneet Arora, and Navpreet Kaloty.
\newblock {iPerceive}: Applying common-sense reasoning to multi-modal dense
  video captioning and video question answering.
\newblock In {\em Proc. WACV}, 2021.

\bibitem{chen2021towards}
Shaoxiang Chen and Yu-Gang Jiang.
\newblock Towards bridging event captioner and sentence localizer for weakly
  supervised dense event captioning.
\newblock In {\em Proc. CVPR}, 2021.

\bibitem{deng2021sketch}
Chaorui Deng, Shizhe Chen, Da Chen, Yuan He, and Qi Wu.
\newblock Sketch, ground, and refine: Top-down dense video captioning.
\newblock In {\em CVPR}, 2021.

\bibitem{deng2020retinaface}
Jiankang Deng, Jia Guo, Evangelos Ververas, Irene Kotsia, and Stefanos
  Zafeiriou.
\newblock {RetinaFace}: Single-shot multi-level face localisation in the wild.
\newblock In {\em Proc. CVPR}, 2020.

\bibitem{dognin2020image}
Pierre Dognin, Igor Melnyk, Youssef Mroueh, Inkit Padhi, Mattia Rigotti, Jarret
  Ross, Yair Schiff, Richard~A Young, and Brian Belgodere.
\newblock Image captioning as an assistive technology: Lessons learned from
  {VizWiz} 2020 challenge.
\newblock {\em arXiv preprint arXiv:2012.11696}, 2020.

\bibitem{Everingham06a}
Mark Everingham, Josef Sivic, and Andrew Zisserman.
\newblock ``{H}ello! {M}y name is... {Buffy}'' -- automatic naming of
  characters in {TV} video.
\newblock In {\em Proc. BMVC}, 2006.

\bibitem{fujita2020soda}
Soichiro Fujita, Tsutomu Hirao, Hidetaka Kamigaito, Manabu Okumura, and Masaaki
  Nagata.
\newblock Soda: Story oriented dense video captioning evaluation framework.
\newblock In {\em Proc. ECCV}, pages 517--531. Springer, 2020.

\bibitem{gurari2020captioning}
Danna Gurari, Yinan Zhao, Meng Zhang, and Nilavra Bhattacharya.
\newblock Captioning images taken by people who are blind.
\newblock In {\em Proc. ECCV}, pages 417--434. Springer, 2020.

\bibitem{Han23}
Tengda Han, Max Bain, Arsha Nagrani, G{\"u}l Varol, Weidi Xie, and Andrew
  Zisserman.
\newblock {AutoAD: Movie Description in Context}.
\newblock In {\em Proc. CVPR}, 2023.

\bibitem{huang2020multimodal}
Gabriel Huang, Bo Pang, Zhenhai Zhu, Clara Rivera, and Radu Soricut.
\newblock Multimodal pretraining for dense video captioning.
\newblock {\em arXiv preprint arXiv:2011.11760}, 2020.

\bibitem{huang2018person}
Qingqiu Huang, Wentao Liu, and Dahua Lin.
\newblock Person search in videos with one portrait through visual and temporal
  links.
\newblock In {\em Proc. ECCV}, pages 425--441, 2018.

\bibitem{huang2020movienet_short}
Qingqiu Huang, Yu Xiong, Anyi Rao, Jiaze Wang, and Dahua Lin.
\newblock {MovieNet}: A holistic dataset for movie understanding.
\newblock In {\em ECCV}, 2020.

\bibitem{huang2020caption}
Qingqiu Huang, Lei Yang, Huaiyi Huang, Tong Wu, and Dahua Lin.
\newblock Caption-supervised face recognition: Training a state-of-the-art face
  model without manual annotation.
\newblock In {\em Proc. ECCV}, 2020.

\bibitem{iashin2020better}
Vladimir Iashin and Esa Rahtu.
\newblock A better use of audio-visual cues: Dense video captioning with
  bi-modal transformer.
\newblock In {\em BMVC}, 2020.

\bibitem{iashin2020multi}
Vladimir Iashin and Esa Rahtu.
\newblock Multi-modal dense video captioning.
\newblock In {\em CVPR Workshops on Multimodal Learning}, 2020.

\bibitem{Korbar22}
Bruno Korbar and Andrew Zisserman.
\newblock Personalised clip or: how to find your vacation videos.
\newblock In {\em British Machine Vision Conference}, 2022.

\bibitem{krishna2017dense}
Ranjay Krishna, Kenji Hata, Frederic Ren, Li Fei-Fei, and Juan Carlos~Niebles.
\newblock Dense-captioning events in videos.
\newblock In {\em Proc. ICCV}, pages 706--715, 2017.

\bibitem{kukleva2020learning}
Anna Kukleva, Makarand Tapaswi, and Ivan Laptev.
\newblock Learning interactions and relationships between movie characters.
\newblock In {\em Proc. CVPR}, pages 9849--9858, 2020.

\bibitem{Laptev08}
Ivan Laptev, Marcin Marsza{\l}ek, Cordelia Schmid, and Benjamin Rozenfeld.
\newblock Learning realistic human actions from movies.
\newblock In {\em Proc. CVPR}, 2008.

\bibitem{lei2018tvqa}
Jie Lei, Licheng Yu, Mohit Bansal, and Tamara~L Berg.
\newblock Tvqa: Localized, compositional video question answering.
\newblock In {\em EMNLP}, 2018.

\bibitem{lei2020tvr}
Jie Lei, Licheng Yu, Tamara~L Berg, and Mohit Bansal.
\newblock {TVR}: A large-scale dataset for video-subtitle moment retrieval.
\newblock In {\em Proc. ECCV}, pages 447--463. Springer, 2020.

\bibitem{li2018jointly}
Yehao Li, Ting Yao, Yingwei Pan, Hongyang Chao, and Tao Mei.
\newblock Jointly localizing and describing events for dense video captioning.
\newblock In {\em Proc. CVPR}, 2018.

\bibitem{lin2004rouge}
Chin-Yew Lin.
\newblock Rouge: A package for automatic evaluation of summaries.
\newblock In {\em Text summarization branches out}, pages 74--81, 2004.

\bibitem{lin2022swinbert}
Kevin Lin, Linjie Li, Chung-Ching Lin, Faisal Ahmed, Zhe Gan, Zicheng Liu,
  Yumao Lu, and Lijuan Wang.
\newblock Swin{BERT}: End-to-end transformers with sparse attention for video
  captioning.
\newblock In {\em Proc. CVPR}, 2022.

\bibitem{loshchilov2017adamw}
Ilya Loshchilov and Frank Hutter.
\newblock Decoupled weight decay regularization.
\newblock {\em arXiv preprint arXiv:1711.05101}, 2017.

\bibitem{luo2020univilm}
Huaishao Luo, Lei Ji, Botian Shi, Haoyang Huang, Nan Duan, Tianrui Li, Xilin
  Chen, and Ming Zhou.
\newblock {UniViLM}: A unified video and language pre-training model for
  multimodal understanding and generation.
\newblock {\em arXiv preprint arXiv:2002.06353}, 2020.

\bibitem{mokady2021clipcap}
Ron Mokady, Amir Hertz, and Amit~H Bermano.
\newblock {ClipCap}: {CLIP} prefix for image captioning.
\newblock {\em arXiv preprint arXiv:2111.09734}, 2021.

\bibitem{mun2019streamlined}
Jonghwan Mun, Linjie Yang, Zhou Ren, Ning Xu, and Bohyung Han.
\newblock Streamlined dense video captioning.
\newblock In {\em Proc. CVPR}, 2019.

\bibitem{nadeau2007survey}
David Nadeau and Satoshi Sekine.
\newblock A survey of named entity recognition and classification.
\newblock {\em Lingvisticae Investigationes}, 30(1):3--26, 2007.

\bibitem{Nagrani17b}
Arsha Nagrani and Andrew Zisserman.
\newblock From benedict cumberbatch to sherlock holmes: Character
  identification in tv series without a script.
\newblock In {\em Proc. BMVC}, 2017.

\bibitem{papalampidi2021movie}
Pinelopi Papalampidi, Frank Keller, and Mirella Lapata.
\newblock Movie summarization via sparse graph construction.
\newblock In {\em Proceedings of the AAAI Conference on Artificial
  Intelligence}, volume~35, pages 13631--13639, 2021.

\bibitem{perego2016gains}
Elisa Perego.
\newblock Gains and losses of watching audio described films for sighted
  viewers.
\newblock {\em Target}, 28(3):424--444, 2016.

\bibitem{clip2021}
Alec Radford, Jong~Wook Kim, Chris Hallacy, Aditya Ramesh, Gabriel Goh,
  Sandhini Agarwal, Girish Sastry, Amanda Askell, Pamela Mishkin, Jack Clark,
  Gretchen Krueger, and Ilya Sutskever.
\newblock Learning transferable visual models from natural language
  supervision.
\newblock In {\em Proc. ICML}, 2021.

\bibitem{radford2022whisper}
Alec Radford, Jong~Wook Kim, Tao Xu, Greg Brockman, Christine McLeavey, and
  Ilya Sutskever.
\newblock Robust speech recognition via large-scale weak supervision.
\newblock {\em OpenAI blog}, 2022.

\bibitem{rahman2019watch}
Tanzila Rahman, Bicheng Xu, and Leonid Sigal.
\newblock Watch, listen and tell: Multi-modal weakly supervised dense event
  captioning.
\newblock In {\em Proc. ICCV}, 2019.

\bibitem{rohrbach2015lsmdc}
Anna Rohrbach, Marcus Rohrbach, Niket Tandon, and Bernt Schiele.
\newblock A dataset for movie description.
\newblock In {\em Proc. CVPR}, 2015.

\bibitem{seo2022end}
Paul~Hongsuck Seo, Arsha Nagrani, Anurag Arnab, and Cordelia Schmid.
\newblock End-to-end generative pretraining for multimodal video captioning.
\newblock In {\em Proc. CVPR}, pages 17959--17968, 2022.

\bibitem{sharma2018cc}
Piyush Sharma, Nan Ding, Sebastian Goodman, and Radu Soricut.
\newblock Conceptual captions: A cleaned, hypernymed, image alt-text dataset
  for automatic image captioning.
\newblock In {\em Association for Computational Linguistics}, 2018.

\bibitem{shen2017weakly}
Zhiqiang Shen, Jianguo Li, Zhou Su, Minjun Li, Yurong Chen, Yu-Gang Jiang, and
  Xiangyang Xue.
\newblock Weakly supervised dense video captioning.
\newblock In {\em Proc. CVPR}, 2017.

\bibitem{shi2019dense}
Botian Shi, Lei Ji, Yaobo Liang, Nan Duan, Peng Chen, Zhendong Niu, and Ming
  Zhou.
\newblock Dense procedure captioning in narrated instructional videos.
\newblock In {\em Association for Computational Linguistics}, 2019.

\bibitem{soldan2022mad}
Mattia Soldan, Alejandro Pardo, Juan~Le{\'o}n Alc{\'a}zar, Fabian Caba, Chen
  Zhao, Silvio Giancola, and Bernard Ghanem.
\newblock {MAD}: A scalable dataset for language grounding in videos from movie
  audio descriptions.
\newblock In {\em Proc. CVPR}, 2022.

\bibitem{Tapaswi12}
Makarand Tapaswi, Martin B{\"a}uml, and Rainer Stiefelhagen.
\newblock ``knock! knock! who is it?'' probabilistic person identification in
  {TV} series.
\newblock In {\em Proc. CVPR}, 2012.

\bibitem{tapaswi2015book2movie}
Makarand Tapaswi, Martin Bauml, and Rainer Stiefelhagen.
\newblock Book2movie: Aligning video scenes with book chapters.
\newblock In {\em Proc. CVPR}, pages 1827--1835, 2015.

\bibitem{tapaswi2019}
Makarand Tapaswi, Marc~T. Law, and Sanja Fidler.
\newblock Video face clustering with unknown number of clusters.
\newblock In {\em Proc. ICCV}, 2019.

\bibitem{vedantam2015cider}
Ramakrishna Vedantam, C Lawrence~Zitnick, and Devi Parikh.
\newblock Cider: Consensus-based image description evaluation.
\newblock In {\em Proc. CVPR}, pages 4566--4575, 2015.

\bibitem{wang2018bidirectional}
Jingwen Wang, Wenhao Jiang, Lin Ma, Wei Liu, and Yong Xu.
\newblock Bidirectional attentive fusion with context gating for dense video
  captioning.
\newblock In {\em Proc. CVPR}, 2018.

\bibitem{wang2021end}
Teng Wang, Ruimao Zhang, Zhichao Lu, Feng Zheng, Ran Cheng, and Ping Luo.
\newblock End-to-end dense video captioning with parallel decoding.
\newblock In {\em Proc. ICCV}, 2021.

\bibitem{wang2020event}
Teng Wang, Huicheng Zheng, Mingjing Yu, Qian Tian, and Haifeng Hu.
\newblock Event-centric hierarchical representation for dense video captioning.
\newblock {\em IEEE Transactions on Circuits and Systems for Video Technology},
  2020.

\bibitem{wang2021toward}
Yujia Wang, Wei Liang, Haikun Huang, Yongqi Zhang, Dingzeyu Li, and Lap-Fai Yu.
\newblock Toward automatic audio description generation for accessible videos.
\newblock In {\em Proceedings of the 2021 CHI Conference on Human Factors in
  Computing Systems}, pages 1--12, 2021.

\bibitem{Xie19a}
Weidi Xie, Arsha Nagrani, Joon~Son Chung, and Andrew Zisserman.
\newblock Utterance-level aggregation for speaker recognition in the wild.
\newblock In {\em Proc. ICASSP}, 2019.

\bibitem{xiong2019graph}
Yu Xiong, Qingqiu Huang, Lingfeng Guo, Hang Zhou, Bolei Zhou, and Dahua Lin.
\newblock A graph-based framework to bridge movies and synopses.
\newblock In {\em Proc. ICCV}, pages 4592--4601, 2019.

\bibitem{yang2023vid2seq}
Antoine Yang, Arsha Nagrani, Paul~Hongsuck Seo, Antoine Miech, Jordi
  Pont-Tuset, Ivan Laptev, Josef Sivic, and Cordelia Schmid.
\newblock Vid2seq: Large-scale pretraining of a visual language model for dense
  video captioning.
\newblock {\em arXiv preprint arXiv:2302.14115}, 2023.

\bibitem{yu2021transitional}
Youngjae Yu, Jiwan Chung, Heeseung Yun, Jongseok Kim, and Gunhee Kim.
\newblock Transitional adaptation of pretrained models for visual storytelling.
\newblock In {\em Proc. CVPR}, pages 12658--12668, 2021.

\bibitem{yu2020character}
Youngjae Yu, Jongseok Kim, Heeseung Yun, Jiwan Chung, and Gunhee Kim.
\newblock Character grounding and re-identification in story of videos and text
  descriptions.
\newblock In {\em Proc. ECCV}, pages 543--559. Springer, 2020.

\bibitem{zhang2019bertscore}
Tianyi Zhang, Varsha Kishore, Felix Wu, Kilian~Q Weinberger, and Yoav Artzi.
\newblock Bertscore: Evaluating text generation with bert.
\newblock In {\em Proc. ICLR}, 2020.

\bibitem{youcook2}
Luowei Zhou, Chenliang Xu, and Jason~J Corso.
\newblock Towards automatic learning of procedures from web instructional
  videos.
\newblock In {\em AAAI}, 2018.

\bibitem{zhou2018end}
Luowei Zhou, Yingbo Zhou, Jason~J Corso, Richard Socher, and Caiming Xiong.
\newblock End-to-end dense video captioning with masked transformer.
\newblock In {\em Proc. CVPR}, 2018.

\bibitem{Zhu2015AligningBA}
Yukun Zhu, Ryan Kiros, Richard~S. Zemel, Ruslan Salakhutdinov, Raquel Urtasun,
  Antonio Torralba, and Sanja Fidler.
\newblock Aligning books and movies: Towards story-like visual explanations by
  watching movies and reading books.
\newblock In {\em Proc. ICCV}, 2015.

\end{thebibliography}
}

\clearpage
\renewcommand{\thefigure}{A.\arabic{figure}} 
\setcounter{figure}{0} 
\renewcommand{\thetable}{A.\arabic{table}}
\setcounter{table}{0} 

\noindent{\Large \textbf{Appendix}}
\appendix

\section{Downloading cast information}
As briefly described in the main paper Section 4.3,
We download cast information from IMDb\footnote{\url{https://www.imdb.com/}}.
Specifically, we first query the movie based on its IMDb ID,~\eg~{tt0120780},
which is provided by the datasets like AudioVault-AD~\cite{Han23} or MAD~\cite{soldan2022mad}.
Next, we download the cast list under the HTML element `\texttt{<span>}Top Cast\texttt{</span>}', where each item in the list contains the actor name, the character name and a portrait picture of the actor.
For each movie, we download such information for up to 10 characters. 

\noindent\paragraph{Special Cases.}
Some characters in the cast list do not have corresponding portrait pictures.
Among 488 movies from MAD-train, 
we find 293 movies have missing portrait pictures in their top-10 cast list.
By manual verification, 
we find it is typically because the actors are less known and therefore do not have an IMDb profile page -- since most of the IMDb data source is contributed by volunteers, there exists an inevitable bias towards celebrities or well-known movies.
In such cases, we remove the characters in our data collection pipeline.
Overall, among 488 movies from MAD-train,
there are 17 movies with less than 5 characters downloaded,
and one movie has an empty character list, which is \textit{Human Flow (2017)}\footnote{\url{https://www.imdb.com/title/tt6573444/}}, a documentary.

\section{Statistics of movie AD and subtitles}

\paragraph{Frequency of names and pronouns.}
Table~\ref{table:appendix:freq_ad} and~\ref{table:appendix:freq_sub} show the frequency of names and pronouns on AD and subtitles respectively.
The frequency is calculated on a per-sentence basis, that is, if any name (from Named-Entity Recognition (NER) outputs) or pronoun exists in the AD/subtitle sentence, the count is accumulated by one. 
The tables show that a substantial 39.1\% of AD sentences contain character names, compared to only 13.3\% for subtitles.
Generating sentences with correct names is an important aspect of AD quality.
Note that in this analysis, 
we discard the intro and outro of the movie for more reliable frequencies. 
The AD during those periods mainly performs an OCR task -- introducing the producers, the name of the studio or reading movie credits at the end, which includes a large number of `[PER]' tags from the NER outputs.

\begin{table}[t]
\centering\small
\setlength{\tabcolsep}{3pt}
\begin{tabular}{lrl}\toprule
from 488 \textbf{MAD-train} movies & quantity & ratio \\ \midrule
all AD sentences & 310,494 & 100\% \\ \midrule
AD with [PER] tag & 121,557 & 39.1\% (40.7\%$^\dagger$) \\
AD with pronouns$^*$ & 111,974 & 36.1\% \\
AD with ([PER] tag \textit{or} pronouns) & 202,256 & 65.1\% \\
\bottomrule
\end{tabular}
\vspace{1mm}
\caption{Frequency of names or pronouns in the {\bf AD sentences}. 
The numbers are based on MAD-train movies \emph{after removing the intro and outro} of the movies. The `[PER]' is the entity category for `person' from NER outputs.
`$\dagger$': If including AD from intro and outro, the percentage of AD with [PER] tag is 40.7\%, which is reported in the main paper page-4 and 8.
`*': We count the occurrence of any one of six pronouns \{she, her, he, him, they, them\}.}\label{table:appendix:freq_ad}\vspace{2mm}
\end{table}

\begin{table}[t]
\centering\small
\begin{tabular}{lrl}\toprule
from 488 \textbf{MAD-train} movies & quantity & ratio \\ \midrule
all subtitle sentences & 628,613 & 100\% \\ \midrule
subtitles with [PER] tag & 83,904 & 13.3\% \\
subtitles with pronouns$^*$ & 150,564 & 24.0\% \\
subtitles with ([PER] tag \textit{or} pronouns) & 216,410 & 34.4\% \\
\bottomrule
\end{tabular}
\vspace{1mm}
\caption{Frequency of names or pronouns in the {\bf subtitles}. 
The numbers are based on MAD-train movies. 
The `[PER]' is the entity category for `person' from NER outputs.
`*': We count the occurrence of any one of eight pronouns \{she, her, he, him, they, them, i, me\}.}\label{table:appendix:freq_sub}
\vspace{-2mm}
\end{table}

\vspace{-2mm}
\paragraph{Unique names within each movie.}
From the NER output of AD sentences, 
we aggregate the unique words with `[PER]' tags for each movie.
For 488 movies in MAD-train, 
we found on average there are 69 unique names for each movie,
with a maximum of 176 unique names and a minimum of 3 unique names.
The number is much higher than the length of a typical cast list because 
(i) characters could be mentioned in different ways,~\eg~by their first-name, last-name or titles, 
(ii) the names mentioned in AD do not correspond to characters,~\eg~Gryffindor for the college name, 
(iii) errors or noises of the NER pipeline that the words are partitioned incorrectly.

\begin{figure*}[h!]
    \centering
    \includegraphics[width=0.99\textwidth]{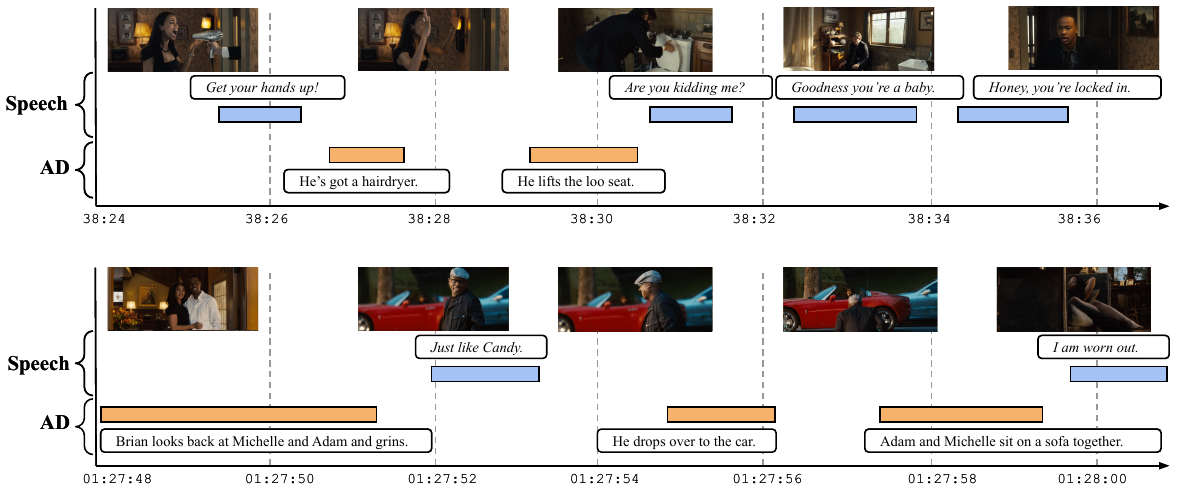}
    \includegraphics[width=0.99\textwidth]{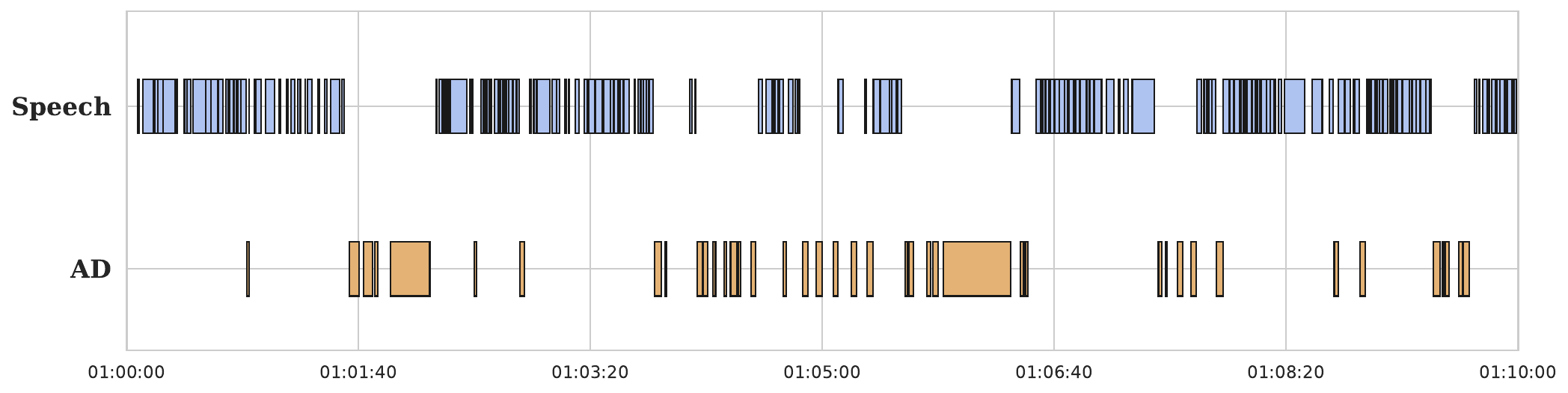}
    \caption{Timeline visualization of a movie with its original dialogue (speech) and human-generated Audio Description (AD). AD is inserted at appropriate times between speech, describing relevant visual elements in the frames. 
    The top and mid figures show movie clips spanning 15 seconds with corresponding frames and texts,
    the bottom figure shows a movie clip spanning 10 minutes with only timestamps.
    The movie shown here is \textit{Death at a Funeral (2010)} with IMDb ID tt1321509. The corresponding AD is sourced from AudioVault-AD (ID 17295).}
    \label{fig:appendix:timeline}
\end{figure*}

\vspace{-2mm}
\paragraph{Visualization of AD and subtitles on the time axis.}
Following The Web Content Accessibility Guidelines 2.0~\cite{caldwell2008web}
(also introduced in the main paper Section 3.3),
successful AD should be added during existing pauses in movie dialogues.
In Figure~\ref{fig:appendix:timeline}, 
we visualize both ground-truth AD and movie subtitles on the timeline for 15-second and 10-minute movie clips to illustrate this interleaved property of ground-truth AD and subtitles.

\begin{figure*}[ht]
    \centering
    \includegraphics[width=0.99\textwidth]{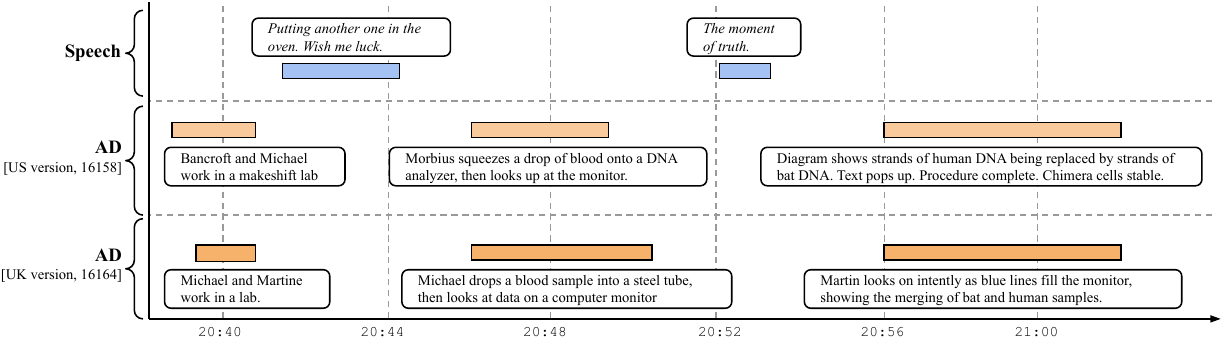}
    \caption{
    Timeline visualization of the \textbf{same movie clip} with its original dialogue (speech) and \textbf{two versions} of human-generated Audio Description (AD).
    Note that disagreements of timestamps exist between different versions of AD for the same movie clip.
    The movie clip is from \textit{Morbius (2022)} with IMDb ID tt5108870.
    The two versions of AD are from AudioVault-AD with ID 16158 (US annotator) and 16164 (UK annotator). 
    The characters who appeared in the scene are \textit{Dr. Michael Morbius} and \textit{Martine Bancroft}.
    }
    \label{fig:appendix:mixed}
\end{figure*}

\vspace{-3mm}
\paragraph{Stats of inter-annotator agreement.}

As briefly described in Section 3.3, the timestamps of human-generated AD vary for the same movie, especially during long pauses in dialogue.
On the AudioVault website, a small portion (less than 20\%) of movies have more than one AD versions or multi-lingual AD versions.
Figure~\ref{fig:appendix:mixed} shows an example movie clip with its two AD versions on AudioVault-AD. Those two versions describe the same movie but are provided by annotators from the US and UK respectively.
Comparing the middle blocks with the lower blocks in Fig.~\ref{fig:appendix:mixed}, it can be seen that AD sentences from the two versions have different start/end timestamps (both shown in {\color{orange} orange} blocks).
We also notice that character names are referred to differently in both AD versions,~\eg the AD at 20:40.
Incorporating multiple versions of AD of the same movie would be an interesting research direction. 
In this paper, we only consider one AD version for each movie by choosing the version with a lower AudioVault ID.

\section{Training details}
\subsection{Character recognition module}
\paragraph{Architecture details.} See Table~\ref{table:appendix:arch_char} for the details of character recognition module.

\begin{table}[h]
\centering\small\vspace{-3mm}
\begin{tabular}{ll}
\toprule
linear projection layer & 512 $\rightarrow$ 512 \\
num blocks & 2 \\
channel & 512 \\
num head & 8 \\
ff dimension & 2048 \\ \bottomrule
\end{tabular}
\vspace{1mm}
\caption{
The architecture details of the character recognition module, which consists of a 2-layer transformer decoder.}\label{table:appendix:arch_char}
\end{table}

\paragraph{Training recipe.}
The character recognition module is trained with binary labels derived from MovieNet face annotations, as described in the main paper Sections 3.2 and 4.1.
The model is trained with AdamW optimizer with a learning rate of $10^{-4}$ for 10 epochs with a batch size of 512 movie clips. The loss is binary cross-entropy with label balancing.

\subsection{Other pretraining with partial data.}
We follow~\cite{Han23} for the pretraining with partial data.
Specifically, we use the text-only AudioVault-AD dataset to finetune the last 6 blocks of a Web-Text pretrained GPT2 for 5 epochs. 
We also use the video-text data from WebVid to pretrain the perceiver resampler and X-Attn blocks for 5 epochs, but with GPT2 weights frozen.
Both pretraining procedures can be achieved in parallel, 
and the trained weights from both settings can be combined as an initialization for the AD generation finetuning.

\subsection{The final finetuning.}
\paragraph{Architecture details.} See Table~\ref{table:appendix:arch_ad} for the details of the perceiver resampler and X-Attn blocks.

\begin{table}[h]
\centering\small
\begin{tabular}{lll}
\toprule
\multirow{6}{*}{Perceiver Resampler} & projection layer$^{\dagger}$ & 512 $\rightarrow$ 768 \\
 & num latent & 10 \\
 & num blocks & 2 \\
 & channel & 768 \\
 & num head & 12 \\
 & ff dimension & 3072 \\ \midrule
\multirow{4}{*}{X-Attn} & num blocks & 12$^{*}$ \\
 & channel & 768 \\
 & num head & 12 \\
 & ff dimension & 3072 \\ \bottomrule
\end{tabular}
\vspace{1mm}
\caption{The architecture details of perceiver resampler and X-Attn blocks.
$^{\dagger}$: The perceiver resampler takes 512-d CLIP visual features as input. Those features are first projected to 768-d for further computation. 
$^{*}$: We insert 12 X-Attn blocks into 12-block GPT2-small model, that is one X-Attn block for each GPT2 block.
}\label{table:appendix:arch_ad}
\end{table}

\vspace{-3mm}
\paragraph{Training recipe.}
The AD generation pipeline is trained (or finetuned) on MAD-train data with a batch size of 64 movie clips for 10 epochs. 
We use the AdamW optimizer with a cosine-decayed learning rate schedule with a linear warm-up. The default learning rate is $10^{-4}$.
The GPT2 weights are frozen when training for AD generation. The trainable parameters are the perceiver resampler and the X-Attn blocks.
For the textual character information (\eg~Jack played by Leonardo DiCaprio ...),
we right-pad the sequences of text tokens for up to 64 tokens.
For the contextual AD information, 
we right-pad the sequences for up to 32 tokens.
For the character's exemplar features, we pad with zero values for up to 10 characters.

\subsection{Temporal Proposal Classification}
\label{app:temp_class}
\paragraph{Architecture Details.} A pretrained BERT \emph{base-uncased} model is used, with special tokens added to the vocabulary for the timestamps tokens, \texttt{|<|t01|>,...,<|t59|>} to indicate each 0.5-second bin in the 30-second context window. The visual CLIP features are first projected through a linear layer (512$\rightarrow$768), whereas the audio features are simply zero-padded from 128$\rightarrow$768. BERT positional embeddings are added to both features.

\begin{figure}[t!]
    \centering
    \includegraphics[width=0.45\textwidth]{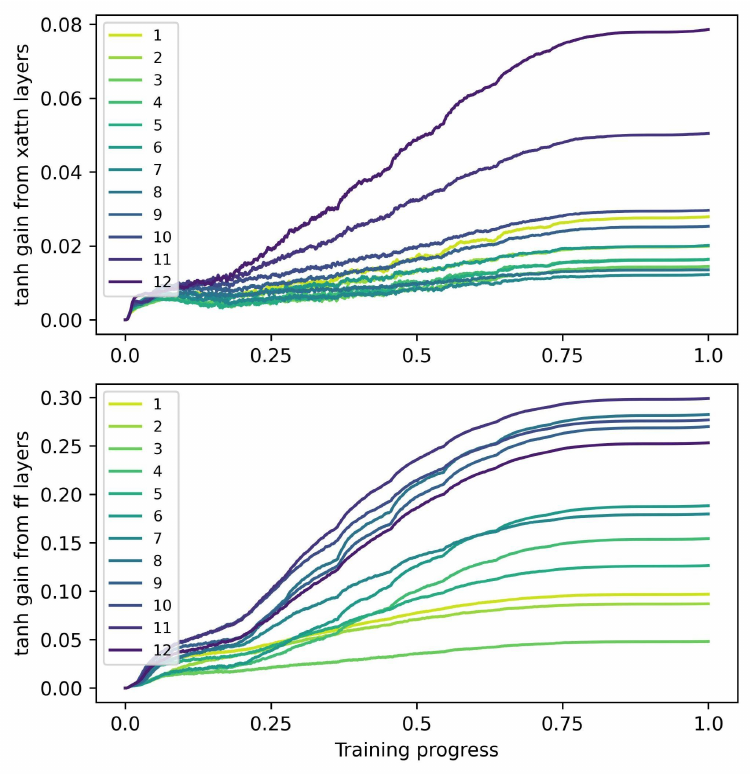}
    \caption{Monitoring Tanh gating during the training process.
    There are two Tanh gates for each X-Attn block: one for X-Attn operation and the other for feed-forward operation. Please refer to~\cite{alayrac2022flamingo} for details. In this figure, the X-Attn blocks are trained from randomly initialized weights, thus the gating value starts from zero.}
    \label{fig:appendix:tanh}
\end{figure}

\begin{figure*}[h!]
    \centering
    \includegraphics[width=0.99\textwidth]{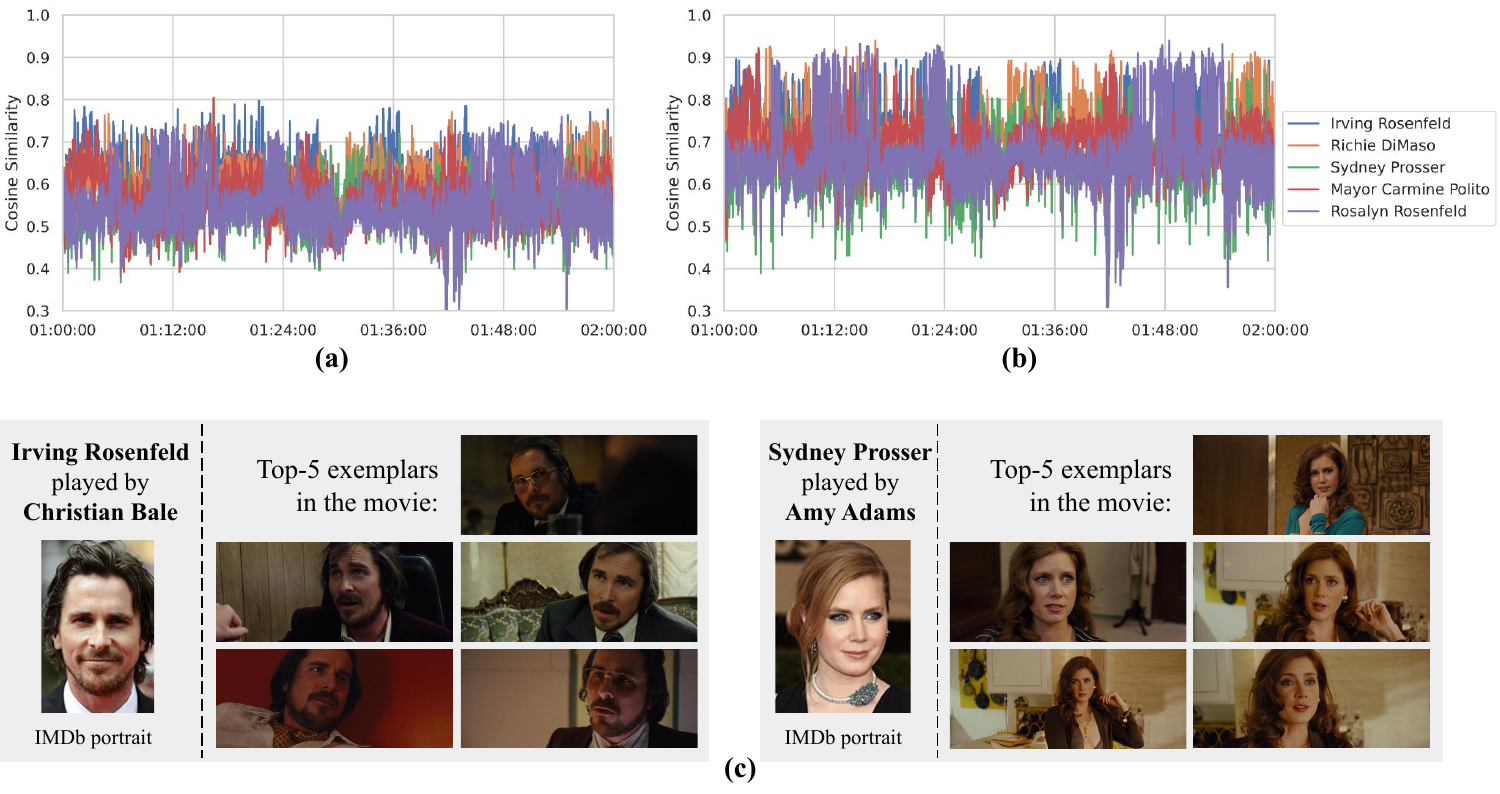}
    \caption{
    Details of calibrating cosine distance and leveraging IMDb portrait images.
    \textbf{(a)} Cosine similarity between actors' IMDb portrait images and the movie features \emph{before} calibration (only a one-hour clip is shown for clarity).
    \textbf{(b)} Cosine similarity between characters' in-movie exemplar features with the movie features,~\ie~\emph{after} calibration. The same one-hour clip is shown.
    \textbf{(c)} Visualization of top-5 exemplars for two characters, which are simply obtained by taking the top-5 peaks from Fig.(a) for each actor.
    The movie samples are from \textit{American Hustle (2013)} with IMDb ID tt1800241.
    }
    \label{fig:appendix:cos_curve}
\end{figure*}

\begin{figure}[t!]
    \centering
    \includegraphics[width=0.45\textwidth]{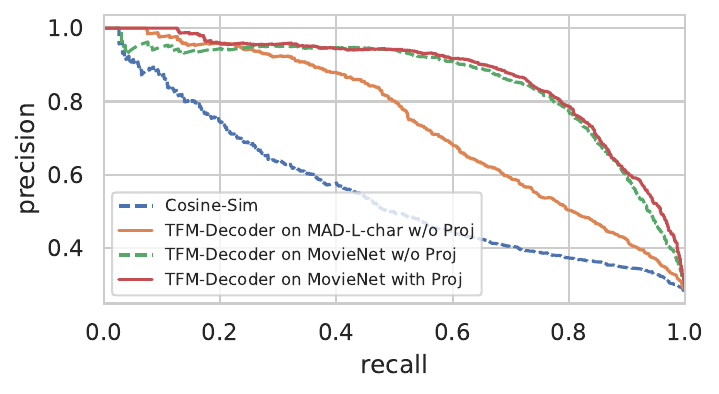}
    \vspace{-1mm}
    \caption{
    More Precision-Recall curves for the character recognition methods.
    {We show three methods: thresholding actor-movie cosine similarity,
    transformer decoder on MAD-L-char w/o linear projection layer, transformer decoder on MovieNet w/o linear projection layer, 
    and transformer decoder on MovieNet with linear projection layer.
    The precision/recall is calculated on a per-character basis,
    ~\ie~the precision/recall of the cosine thresholds to correctly find a character name mentioned in the AD. }}
    \label{fig:appendix:pr_curve}
\end{figure}

\paragraph{Training recipe.}
The model is trained with a batch size of 64 context windows for 3 epochs on MAD-train movies. We use AdamW optimizer with a learning rate of $10^{-4}$.

\paragraph{Baseline.}
For the binary temporal proposal classification task described in Section~\ref{method:time}, we propose a simple decision-based baseline whereby any speech gap with a duration greater than a fixed threshold is classified to have AD inserted, and not AD inserted otherwise. 
In Table~\ref{tab:time_proposal}, the Average Precision and ROC AUC is calculated by varying the fixed threshold at 100 values equally spaced between 2 and 6 seconds.

\section{Analysis}
\subsection{Tanh gating during training}
Following Flamingo~\cite{alayrac2022flamingo},
we visualize the absolute value of tanh gating for each X-Attn block during training,
which could be a rough indicator showing how much visual information is conditioned by the GPT-2 model.
In contrast to Flamingo Fig. 6 that their tanh gating values are much closer to 1,
our Figure~\ref{fig:appendix:tanh} shows the tanh values have a similar increasing trend during training but the final value is much lower. 
It indicates a longer training schedule with a larger dataset would further benefit our model.

\begin{figure*}[t!]
    \centering
    \includegraphics[width=0.99\textwidth]{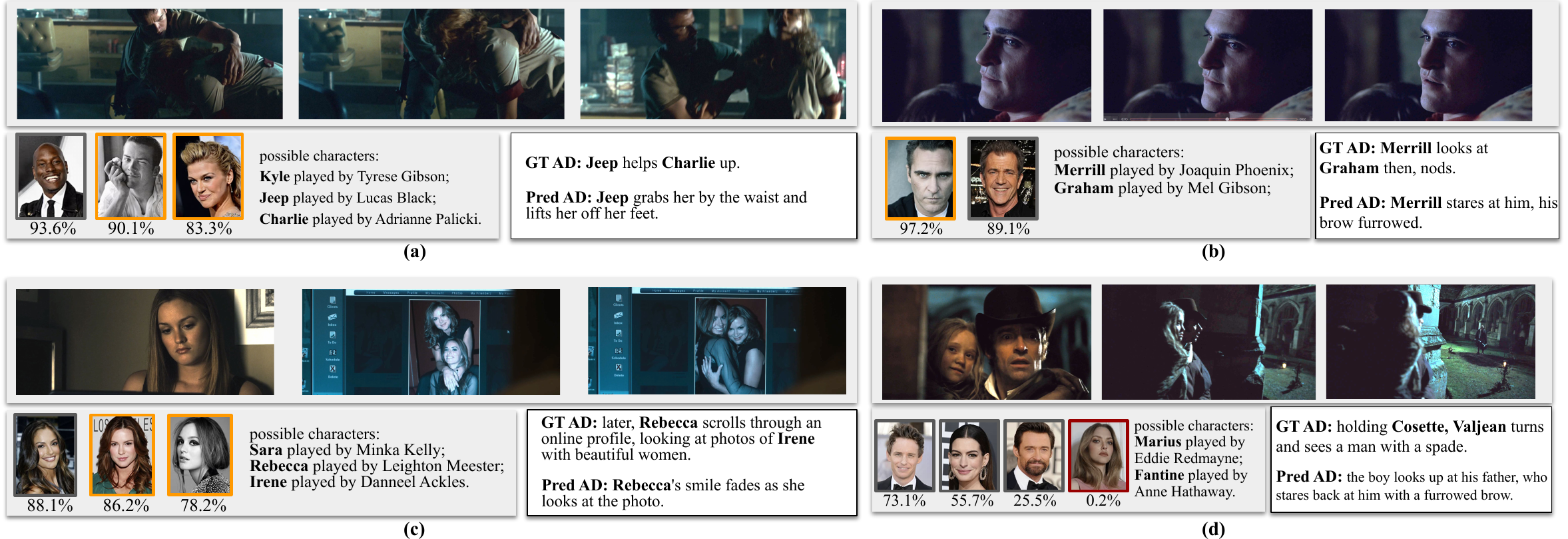}
    \caption{
    Following the same style as the main paper Figure~\ref{fig:qualitative},
    we show qualitative results with the character bank.
    The probability shown below the characters’ portraits is the output of our character recognition module, with correctly recognized characters marked using an {\color{orange} orange} border.
    We use 50\% as the decision boundary for active characters. 
    The movies are from
    \textbf{(a)}: Legion (2010), 
    \textbf{(b)}: Signs (2002), 
    \textbf{(c)}: The Roommate (2011), 
    \textbf{(d)}: Les Mis\'erables (2012).}\label{fig:appendix:qual}
    \vspace{-1mm}
\end{figure*}

\subsection{Character recognition module}
\paragraph{Cosine distance and calibration.}
As shown in Figure~\ref{fig:appendix:cos_curve}-(a),
the cosine similarity between actors' IMDb portrait images and the movie features is not a good indicator of in-screen or off-screen actors. For example, the peaks of the blue curve (Irving Rosenfeld) are always higher than that of the purple curve (Rosalyn Rosenfeld).
As introduced in the main paper page 4, in order to compensate for the variance of appearance from IMDb portrait images, we find exemplars of the actors in the same movie as a calibration process. 
Figure~\ref{fig:appendix:cos_curve}-(c) shows two examples of exemplar searching, which is achieved by simply taking the top-5 peaks for each actor in Fig.~\ref{fig:appendix:cos_curve}-(a).
Next, we use the averaged exemplar features to replace the original IMDb portrait features and re-compute the cosine similarity.
As shown in Figure~\ref{fig:appendix:cos_curve}-(b),
the calibration process normalizes the cosine similarity and makes the comparison between actors more meaningful.

\paragraph{Other character annotation dataset.}
In the main paper, 
we use the manually annotated character annotation from the MovieNet dataset.
But the character labels can also be obtained with weakly annotated data, such as the AD annotation.

We propose a dataset named \textbf{MAD-L-char} for movie character recognition,
which is sourced from MAD-train and LSMDC-train.
The character names in \textbf{MAD-L-char} are automatically mined in two steps: (1) running named entity recognition (NER)~\cite{nadeau2007survey} on the AD annotation, and (2) computing the intersection with the movie's cast list.
Specifically, the NER on MAD-train is sourced by running an open-sourced model\footnote{https://huggingface.co/Jean-Baptiste/camembert-ner}, and the NER from 139 LSMDC-train movies can be obtained from the LSMDC annotations.

\vspace{-1mm}
\paragraph{P-R curve for character recognition.}

In addition to the main paper Table 1 and Fig. 5,
here in Fig.~\ref{fig:appendix:pr_curve},
we compare four PR curves as detailed in the figure caption.
The PR curves show that the model trained on the manually annotated MovieNet dataset clearly outperforms the same model trained on the automatically mined MAD-L-char dataset. Additionally, the extra linear project layer brings a clear performance gain.
{
Note that it is difficult to achieve perfect PR curves, 
partially because for some movies, even the top 10 characters downloaded from IMDb may not cover the main characters, such as the Harry Potter series which has a very large cast list. 
}
The corresponding quantitative metrics of these methods are shown in Table~\ref{table:appendix:pr}.

\begin{table}[t]
    \footnotesize
    \setlength{\tabcolsep}{2pt}
    \centering
    \begin{tabular}{ccccc}
    \toprule
    Methods        & Training Data & Linear Proj  & ROC AUC       & Average Precision \\ \midrule
    Cosine-Sim     & - & -                 & {0.72}          &  {0.55}             \\
    TFM Decoder    & MAD-L-char & \xmark     & {0.84} & {0.74}     \\
    TFM Decoder    & MovieNet & \xmark     & {0.92} & {0.85}    \\
    TFM Decoder    & MovieNet & \cmark     & {\textbf{0.93}} & {\textbf{0.87}}     \\
    \bottomrule
    \end{tabular}
    \vspace{1mm}
    \caption{Quantitative comparison of various character recognition modules.}
    \label{table:appendix:pr}
    \vspace{-3mm}
\end{table}

\vspace{-2mm}
\paragraph{Statistics of recognized active characters.}
After the character recognition module is trained, 
we simply choose the standard probability of 0.5 as the threshold for the decision boundary.
{
With a threshold of 0.5, the character recognition module achieves 0.83 recall and 0.75 precision on MAD-eval movies (read from Figure~\ref{fig:appendix:pr_curve}).
Next, this module can be used to recognize active characters in any public movie, either offline or on-the-fly.
Among more than 300k AD sentences in MAD-train, 
the character recognition module predicts 1.3 active characters on average per AD sentence, 
with 94.8\% AD sentences having no more than 5 predicted active characters
and 14.6\% AD sentences having zero active characters.
}

\subsection{Learning with subtitles}
In addition to the character bank, we find feeding in subtitles as model inputs does not further improve performance. There are two possible reasons: 
(i) usually the subtitles do not describe the scene or characters, and
(ii) the character names are already supplied by the character bank.
Leveraging movie subtitles effectively is a promising future direction. 

\section{More qualitative results}
More qualitative results are shown in Figure~\ref{fig:appendix:qual}.
Note that in \textbf{(d)}, the girl in the scene (\emph{young} Cosette played by Isabelle Allen) is not in the top cast whereas our top cast contains the \emph{adult} Cosette played by Amanda Seyfried, shown in {\color{red} red} border.
Recognizing characters in such cases is challenging but it indicates the character recognition module has a large space for improvement.

\begin{figure}[h]
    \centering
    \includegraphics[width=0.45\textwidth]{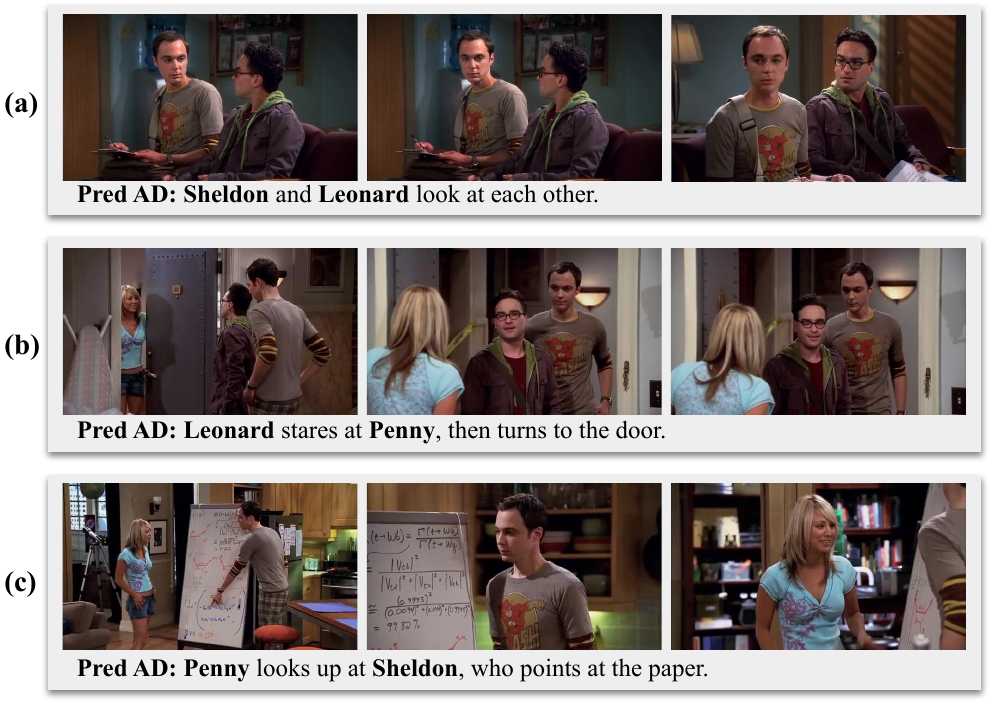}
    \caption{Qualitative results on TVC samples without any specific training. The characters' portraits are downloaded from IMDb page of \textit{The Big Bang Theory}~\url{https://www.imdb.com/title/tt0898266/}.}
    \label{fig:appendix:tvc}
\end{figure}

{
\section{Video captioning results on TVC}
TVC~\cite{lei2020tvr} is a video captioning dataset consisting of TV series, which contains character names in captions.
There are some domain gaps between TV series and movies:
\eg~the frequent character bank is smaller for TV series, and scene locations may be less varied. 
In Figure~\ref{fig:appendix:tvc}, we provide qualitative results of adapting our AutoAD-II model on TVC \emph{without} any further training on TV series. Different from TVC which provides captions for a relatively long video clip spanning a few minutes, we feed in short clips spanning just a few seconds to match our training distribution.
}

\end{document}